\documentclass{article}

\usepackage[main,final]{neurips_2026}
\usepackage{amsmath,amsfonts,bm}

\def\eqref#1{equation~\ref{#1}}
\def\1{\bm{1}}

\DeclareMathAlphabet{\mathsfit}{\encodingdefault}{\sfdefault}{m}{sl}
\SetMathAlphabet{\mathsfit}{bold}{\encodingdefault}{\sfdefault}{bx}{n}

\DeclareMathOperator*{\argmax}{arg\,max}
\DeclareMathOperator*{\argmin}{arg\,min}

\usepackage{url}
\usepackage{booktabs}       % professional-quality tables
\usepackage{amsfonts}       % blackboard math symbols
\usepackage{nicefrac}       % compact symbols for 1/2, etc.
\usepackage{microtype}      % microtypography
\usepackage{xcolor}         % colors
\usepackage{amsmath}
\usepackage{amssymb}
\usepackage{mathtools}
\usepackage{amsthm}
\usepackage{bbm}
\usepackage{multirow}
\usepackage{makecell}
\usepackage{adjustbox}
\usepackage{enumitem}
\usepackage{xcolor}
\usepackage{comment}
\usepackage[linktoc=all]{hyperref}
\usepackage{etoc}

\usepackage{siunitx}  % For aligning numbers by the decimal point (optional but good)
\usepackage{listings}

\usepackage[ruled]{algorithm2e}
\usepackage{tcolorbox}
\usepackage{subcaption}
\usepackage{wrapfig}
\usepackage{graphicx} % For including figures
\usepackage{caption}  % For customizing captions
\tcbuselibrary{skins, breakable, theorems}
\newtcolorbox{conclusionbox}{%
  enhanced, % Use enhanced drawing capabilities
  fonttitle=\bfseries, % Bold title font
  title=Regularization Characterization, % Set the title
  boxrule=0.5pt, % Frame thickness
  arc=4pt, % Rounded corners
  breakable, % Allow the box to break across pages if needed
  left=5pt, % Left padding
  right=5pt, % Right padding
  top=5pt, % Top padding
  bottom=5pt, % Bottom padding
  before skip=5pt, % Space before the box
  after skip=5pt, % Space after the box
}

\hypersetup{
    colorlinks,
    linkcolor={red!50!black},
    citecolor={brown!85!red},
    urlcolor={blue!80!black}
}

\definecolor{codegreen}{rgb}{0,0.6,0}
\definecolor{codegray}{rgb}{0.5,0.5,0.5}
\definecolor{codepurple}{rgb}{0.58,0,0.82}
\definecolor{backcolour}{rgb}{0.95,0.95,0.92}
\definecolor{bestblue}{RGB}{221,235,247}

\lstdefinestyle{mystyle}{
    backgroundcolor=\color{backcolour},   
    commentstyle=\color{codegreen},
    keywordstyle=\color{magenta},
    numberstyle=\tiny\color{codegray},
    stringstyle=\color{codepurple},
    basicstyle=\ttfamily\footnotesize,
    breakatwhitespace=false,         
    breaklines=true,                 
    captionpos=b,                    
    keepspaces=true,                 
    numbers=left,                    
    numbersep=5pt,                  
    showspaces=false,                
    showstringspaces=false,
    showtabs=false,                  
    tabsize=2
}

\usepackage[capitalize,noabbrev]{cleveref}

\usepackage{xcolor,colortbl}

\definecolor{Gray}{gray}{0.85}
\definecolor{LightCyan}{rgb}{0.88,1,1}
\newcolumntype{a}{>{\columncolor{Gray}}c}
\newcolumntype{b}{>{\columncolor{white}}c}

\theoremstyle{plain}
\newtheorem{theorem}{Theorem} % Numbered globally
\newtheorem{lemma}[theorem]{Lemma}     % Numbered globally
\newtheorem{corollary}[theorem]{Corollary}

\theoremstyle{definition}
\newtheorem{definition}[theorem]{Definition}

\theoremstyle{remark}

\newcommand{\AdaBracket}[1]{\left(#1\right)}
\newcommand{\AdaRectBracket}[1]{\left[#1\right]}

\newcommand{\AdaCurlyBracket}[1]{\left\{ #1 \right\}}

\newcommand{\chiAny}[3]{\chi^{#1}\AdaBracket{#2||#3}}
\newcommand{\expectation}[2]{\mathbb{E}_{#1}\AdaRectBracket{#2}}

\newcommand{\fdiv}{$f$-divergence }

\newcommand{\KLany}[2]{D_\text{KL}\!\left(#1 \left|  \right| #2 \right)}

\newcommand{\fdivAny}[2]{D_f\!\left(#1 \left|  \right| #2 \right)}

\newcommand{\datasetPolicy}{\pi_{\mathcal{D}}}

\newcommand{\mynorm}[2]{\left|\!\left| #1 \right|\!\right|_{#2}}
\newcommand{\policydot}[1]{#1(\cdot|s)}
\newcommand{\adaAbsolute}[1]{\left| #1 \right|}
\newcommand{\regD}{D_\text{Reg}}
\newcommand{\regPi}{\pi^*}
\newcommand{\lossD}{D_\text{Opt}}
\newcommand{\lossPi}{\pi_\text{Loss}}

\newcommand{\regN}{N_\text{Reg}}
\newcommand{\lossN}{N_\text{Loss}}

\usepackage[utf8]{inputenc} % allow utf-8 input
\usepackage[T1]{fontenc}    % use 8-bit T1 fonts
\usepackage{hyperref}       % hyperlinks
\usepackage{url}            % simple URL typesetting
\usepackage{booktabs}       % professional-quality tables
\usepackage{amsfonts}       % blackboard math symbols
\usepackage{nicefrac}       % compact symbols for 1/2, etc.
\usepackage{microtype}      % microtypography
\usepackage{xcolor}         % colors

\title{Symmetric Behavior Regularized Policy Optimization}

\author{%
  Lingwei Zhu\thanks{Corresponding Author} \\
  SCIT, Great Bay University\\
  \texttt{zhulingwei@gbu.edu.cn} \\
  \And
  Haseeb Shah \\
  University of Alberta \\
  \texttt{hshah1@ualberta.ca} \\
  \AND
  Zheng Chen \\
  SANKEN, Osaka University \\
  \texttt{chenzcha@gmail.com} \\
  \And
  Martha White \\
  University of Alberta\\
  \texttt{whitem@ualberta.ca} \\
}

\begin{document}

\maketitle

\begin{abstract}

Behavior Regularized Policy Optimization (BRPO) leverages asymmetric divergence regularization to mitigate distribution shift in offline reinforcement learning. 
This paper is the first to study the open question of symmetric BRPO. 
Using didactic examples, we show that symmetric regularization can outperform asymmetric regularization in addressing one-sided bias, near-boundary policy updates, and projection geometry consistency. However, symmetric divergences do not fit BRPO naturally:
they do not permit a closed-form solution when used as regularizers,
and can lead to numerical instability when used as optimization objectives.
We first introduce a universal BRPO framework using an infinite series of Pearson-Vajda divergences to  represent any $f$-divergence, which includes both symmetric and asymmetric divergences.
We use a finite-series approximation to obtain the following results for symmetric BRPO:
(1) a closed-form optimal policy expression;
(2) a numerically stable optimization surrogate; and
(3) a tight upper bound on the approximation quality.
On the D4RL benchmark and in didactic examples, we show that the proposed method achieves consistently strong results and is robust to the number of terms in the approximation.
% MArtha: mostly a non-statement
%opening up possibilities for more diverse and effective regularization choices for offline RL.

\end{abstract}

\newcommand{\eqtagblue}[1]{%
  \makebox[0pt][r]{\text{\colorbox{bestblue}{\scriptsize\strut #1}}\hspace{0.8em}}%
}

\section{Introduction}

% Behavior regularized policy optimization (BRPO) is a simple, yet effective method that has attracted significant research interest for offline reinforcement learning (RL).
% By regularizing towards the behavior policy via a divergence penalty, BRPO effectively suppresses distributional shift incurred by out-of-distribution actions that may have spuriously high values.
% While many works have studied the sensitivity of BRPO to the behavior policy \citep{Kostrikov2022-implicitQlearning,Xiao2025-iterativeRefineCPI}, this paper investigates an orthogonal direction by focusing on the role played by distinct divergence regularizers.
Behavior regularized policy optimization (BRPO) is a simple yet highly effective method for offline reinforcement learning.
By regularizing toward the behavior policy (dataset policy), BRPO reduces the distributional shift induced by out-of-distribution actions.
The choice of divergence for the regularization determines how the distributional shift is controlled, as well as the form of the solution, the projection geometry, and the trade-off between the behavior policy and the action value.
BRPO can be cast as the following two steps:
\begin{align}
   \eqtagblue{\textbf{Regularization}}
   & \regPi = \argmax_{\pi} \,\, \expectation{\substack{s\sim\mathcal{D}\\a\sim\pi}}{
      Q(s,a) - \tau {\color{black}\regD}\!\AdaBracket{{\pi(\cdot | s)} \,||\,{\datasetPolicy(\cdot | s)}}
   }, \label{eq:brpo} \\
   \eqtagblue{\textbf{Optimization}}
   & \pi_{\theta} = \argmin_{\theta} \,\, \expectation{s\sim\mathcal{D}}{
      {\color{black}\lossD}\!\AdaBracket{{\regPi(\cdot | s)} \,||\, {\pi_{\theta}(\cdot | s)}}
   } \label{eq:brpo_optimize}.
\end{align}
% \begin{align}
%    {\color{bestblue}\text{Regularization}} \quad &  \regPi =  \argmax_{\pi} \,\, \expectation{\substack{s\sim\mathcal{D}\\a\sim\pi}}{ Q(s,a) - \tau {\color{black} \regD}\!\AdaBracket{{\pi(\cdot | s)} \,||\,{\datasetPolicy(\cdot | s)}}},  \label{eq:brpo} \\
%    &  \pi_{\theta} = \argmin_{\theta}  \,\, \expectation{s\sim\mathcal{D}}{{\color{black} \lossD}\!\AdaBracket{{{ \regPi (\cdot | s) \,||\, \pi_{\theta}(\cdot | s)} } }} \label{eq:brpo_optimize}.
% \end{align}
 where $\mathcal{D}$ denotes the dataset, $Q$ the action-values, $\regPi$ the regularized optimal policy, $\datasetPolicy$ the behavior policy, and $\pi_{\theta}$ a parameterized policy to be optimized.
 Computing $\regPi$ and sampling from it are usually intractable.
 % Here, Eq. (\ref{eq:brpo}) defines the theoretical maximizer of the regularized objective.
 % However, this maximizer is not practical since computing $\pi^*$ is generally intractable. 
Therefore, the approximation step in Eq. (\ref{eq:brpo_optimize}) is required. 

% Martha: new paragraph here with a topic sentence
A key question for BRPO is how to select $\regD$ and $\lossD$.
The most popular candidate for both is the KL divergence, whose properties have been well studied  \citep{Wu2020-BehaviorRegularizedAC,Jaques2020-OfflineDialog,chan2021-greedification}.
Other candidates include the $\alpha$-divergence and Tsallis divergence \citep{Xu2023-OfflineNoODDAlphaDiv,Zhu2023-tsallisOffline}.
These divergences are all asymmetric: $D_\text{Reg}(\pi||\mu) \neq D_\text{Reg}(\mu || \pi)$ for policies $\pi, \neq \mu$. 
Recently, researchers have noted that symmetric divergences used as the optimization objective $\lossD$ can improve agent performance in LLMs \citep{Go2023-AligningLanguageModels-fDivergence,han-2025-fPO,Li-2026-ChoiceDivergence-RLVROnline}, calling for a more in-depth analysis of symmetric divergences.

 % Many recent works use symmetric D in eq.2 but keep the target as the KL-regularized unchanged \citep{han-2025-fPO,Li-2026-ChoiceDivergence-RLVROnline}. We will show this is suboptimal in section \ref{sec:why_symm}.

This paper derives BRPO with symmetric divergences and motivates the utility for reinforcement learning.
In didactic examples, we first show that 
symmetric divergences can outperform asymmetric ones, due to better handling of support gaps and bias (Section \ref{sec:symm_reg}), and better policy behavior near the action boundary under action clipping (Section \ref{sec:symm_opt}). We additionally motivate the utility of having the same symmetric divergence for both $\regD$ and $\lossD$. Symmetric divergences have in-fact been used in RL, but only for $\lossD$; in these instances, the asymmetric KL divergence was still used for $\regD$ \citep{Go2023-AligningLanguageModels-fDivergence,Li-2026-ChoiceDivergence-RLVROnline}. We show that this mismatch can be suboptimal due to differences in projection geometry (Section \ref{sec:symm_match}). This mismatch, however, was unavoidable, as to date, there has been no proposed way to use a symmetric divergence for both. 
% haseeb: this is redundant: Section \ref{sec:why_symm} is devoted to illustrating these points using didactic examples.

Deriving symmetric BRPO, however, is not straightforward. It is challenging because using symmetric divergences, such as Jensen-Shannon and Jeffreys, for $\regD$ do not permit a closed-form optimal policy $\regPi$ (Section \ref{sec:sbrpo-challenges}).
Therefore, even specifying the optimization objective in Equation \ref{eq:brpo_optimize} becomes challenging.
Moreover, directly optimizing a symmetric $\lossD$, such as Jensen-Shannon, can incur numerical issues, as reported in recent studies \citep{Go2023-AligningLanguageModels-fDivergence}.
We resolve these issues with a \emph{novel universal BRPO framework that can recover all $f$-divergence regularization}, based on the infinite series of Pearson-Vajda divergences.
We prove the following in Section \ref{sec:taylor}.
\begin{enumerate}[leftmargin=15pt]
    \item A closed-form optimal policy can be obtained when the Pearson-Vajda series is truncated to a finite number of terms (Theorem \ref{thm:reg_solution}).  The resulting optimal policy filters low-value actions, thereby effectively controlling out-of-distribution actions and improving learning efficiency.
   \item An improved symmetric optimization surrogate objective that avoids numerical instability and thereby improves performance (Theorem \ref{thm:cond_symm}).
    \item A tight upper bound between the exact symmetric divergence and its approximation, ensuring that the surrogates closely resemble the exact counterparts (Theorem \ref{thm:error_bound}).  
\end{enumerate}
We introduce Symmetric $f$-Actor-Critic (S$f$-AC) and show it outperforms several standard offline RL algorithms on D4RL.
%the  D4RL locomotion/Adroit/Maze benchmarks.
We also investigate the impact of the numbers of terms used in the finite approximation, and verify that beyond the didactic examples, symmetric BRPO behaves better at action boundaries in D4RL.

\section{Preliminaries}

% We introduce basics of offline RL.
% Then the $f$-divergence family encompassing all divergences considered in the paper is introduced.

% \subsection{Offline Reinforcement Learning}

\textbf{Offline RL.} 
We focus on discounted Markov Decision Processes (MDPs) defined by the tuple $(\mathcal{S}, \mathcal{A}, P,  r,  \gamma)$, where $\mathcal{S}$ and $\mathcal{A}$ denote the state and action spaces, respectively. 
$P$ denotes the transition probability,
% Let $\Delta(\mathcal{X})$ denote the set of probability distributions over $\mathcal{X}$.
% $d\in\Delta(\mathcal{S})$ denotes the initial state distribution.
% $P: \mathcal{S}\times\mathcal{A} \rightarrow \Delta(\mathcal{S})$ denotes the transition probability function,  
$r$ defines the reward function,
and $\gamma \in (0,1)$ is the discount factor.
A policy $\pi$ is a mapping from states to distributions over actions.
% The quality of a policy is assessed by the expected return $J(\pi) = \int_{\mathcal{S}} \rho^{\pi}(s) \int_{\mathcal{A}} \pi(a|s) r(s,a) \mathrm{d}a \mathrm{d}s$,
% where $\rho^{\pi}_s = \sum_{t=0}^{\infty} \gamma^t P(s_t=s)$ is the unnormalized state visitation frequency.
% The goal is to learn a policy that maximizes $J(\pi)$.
We define the action-value and state-value functions as $Q^{\pi}(s,a) = \expectation{\pi}{\sum_{t=0}^{\infty}\gamma^t r(s_t, a_t) | s_0 = s, a_0 = a}$ and $V^{\pi}(s) = \expectation{\pi}{Q^{\pi}(s,a)}$.
In this paper, we focus on the offline RL setting, where the agent learns from a static dataset $\mathcal{D} = \{s_i, a_i, r_i, s'_{i}\}$ of transitions.
We denote by $\datasetPolicy$ the behavior policy that generated the dataset $\mathcal{D}$.

Behavior regularized policy optimization (BRPO) for offline RL uses the following objective:
\begin{align}
   % \regPi(a|s) =  \argmax_{\pi} 
   \max_{\pi} \,\,
   \expectation{\substack{s\sim\mathcal{D}\\a\sim\pi}}{Q(s,a) - \tau {\color{black} \regD \!\AdaBracket{{\pi(\cdot | s)} ||{\datasetPolicy (\cdot | s)}}}}, 
   \label{eq:brpo_original}
\end{align}
Eq. (\ref{eq:brpo_original}) is intractable for high-dimensional continuous action spaces because it entails solving the integral over actions.
Therefore, Eq.(\ref{eq:brpo_original}) is broken down into Eqs.(\ref{eq:brpo}), (\ref{eq:brpo_optimize}) in the introduction.
Although we can determine the functional form of $\regPi$ from the convexity of $\regD$ \citep{hiriart2004-convexanalysis}, sampling from it is generally intractable due to the normalization constant.
Instead, we compute a tractable surrogate policy $\pi_{\theta}$ from $\regPi$ by minimizing $\lossD$.  
This surrogate policy is an approximate maximizer of the original objective.

% \subsection{$f$-Divergences and Asymmetric BRPO}

\textbf{$f$-divergence Policy Optimization. }
A $f$-divergence from policy $\mu$ to $\pi$ is defined by:
% \begin{definition}\label{def:fdiv}
    % The $f$-divergence from policy $\mu$ to $\pi$ is defined by:
    \begin{align*}
        \fdivAny{\pi(\cdot|s)}{\mu(\cdot|s)} := \int_{\mathcal{A}}  \mu(a|s)f\!\AdaBracket{\frac{\pi(a|s)}{\mu(a|s)}} \mathrm{d}a =  \expectation{\mu}{f\!\AdaBracket{\frac{\pi(a|s)}{\mu(a|s)}}},
        % \vspace{-1em}
    \end{align*}
where $f$ is a convex, lower semi-continuous function satisfying $f(1)=0$ and we have assumed that $\pi$ is absolutely continuous with respect to $\mu$.
The most popular example is the KL divergence with $f(t) = t\ln t$. In this work, we particularly care about symmetric divergences, as in Definition \ref{def:wider_symm}. The main divergences of interest are Jensen-Shannon (JS), Jeffreys, and a modified JS divergence used in Generative Adversarial Nets (GANs), listed in Table \ref{tab:f-divergences}.\footnote{\citet{Pardo2006-divergences} give an alternative definition, defining symmetric divergences $f(t) + t f\AdaBracket{\frac{1}{t}}$. In machine learning, it is more common to use the more flexible one given by Sason \& Verd\'u' \citep{Sason2016-fDiverInequalities,Nowozin2016-fGAN}.}

\begin{definition}[\citet{Sason2016-fDiverInequalities}]\label{def:wider_symm}
    Symmetric divergences are defined by $f(t) = t\ln t + g(t)$, where $g$ is an arbitrary convex function that we call the \emph{conditional symmetry function}.
\end{definition}
\vspace{-0.5mm}

\begin{table*}[t!]
    \centering
    \resizebox{.99\textwidth}{!}{
    \begin{tabular}{p{2cm}p{2.5cm}p{7cm}p{4cm}}
    \toprule

\multirow{6}{*}{Asymmetric}
 & Divergence & $D_f(\pi||\mu)$ & $f(t)$    \\ 
    \midrule[0.5pt]
      & Forward KL  & $\int \pi(x) \ln \frac{\pi(x)}{\mu(x)} \,\textrm{d}x$  & $t\ln t$ \\ [3pt]
    \cmidrule{2-4}
    % Forward & $-\ln t$ & $(-1)^q (q-1)!$ &  $\sum_{q=2}^{\infty} \frac{(-1)^{q}(q-1)}{q}$ \\ 
    % |rule
    % Backward $\KLindex{k}{k+1}$ & $t\ln t$ & $(-1)^q (q-2)!$  & $\sum_{q=2}^{\infty} \frac{(-1)^{q}}{q}$ \\ 
    % |rule
    &Reverse KL & $\int \mu(x) \ln \frac{\mu(x)}{\pi(x)}\,\textrm{d}x$ 
 &   $-\ln t$  \\ [3pt]
    \midrule[0.5pt]
    \multirow{5}{*}{Symmetric}
    &Jeffreys & $\int\left(\pi(x)-\mu(x)\right) \ln \left(\frac{\pi(x)}{\mu(x)}\right) \,\textrm{d}x$  &  $(t-1) \ln t$ \\ [3pt]
% $\sum_{q=2}^{\infty} \! \frac{(-1)^q (q-1)}{q} \! \AdaBracket{1 + \frac{1}{q-1}}$
    \cmidrule{2-4}
    &Jensen-Shannon   &   
    {\footnotesize $\frac{1}{2} \int \pi(x) \ln \frac{2 \pi(x)}{\pi(x)+\mu(x)}
  + \mu(x) \ln \frac{2 \mu(x)}{\pi(x) + \mu(x)}\,\textrm{d}x$ }
  & $t\ln t - (1+t)\ln\frac{1+t}{2}$  \\ [3pt]
    \cmidrule{2-4}
    &GAN divergence  & {\footnotesize $\int \pi(x) \ln \frac{2 \pi(x)}{\pi(x)+\mu(x)}
  + \mu(x) \ln \frac{2 \mu(x)}{\pi(x) + \mu(x)}\,\textrm{d}x - \ln(4)$}  & $t\ln t - (1+t)\ln\AdaBracket{1+t}$ \\ [3pt]
    \bottomrule
    \end{tabular}}
    \caption{
      Asymmetric and symmetric $f$-divergences, their definitions and the generator functions. 
      Jeffreys divergence equals forward KL plus reverse KL.
      GAN divergence refers to the modified Jensen-Shannon divergence used by Generative Adversarial Nets \citep{GAN}. 
      }
      \label{tab:f-divergences}
      \vspace{-1em}
  \end{table*}

\section{Symmetric Divergences in Regularization and Optimization}\label{sec:why_symm}

This section motivates symmetric divergences using didactic examples.
Sections \ref{sec:symm_reg} and \ref{sec:symm_opt} focus on regularization and optimization, respectively.
Section \ref{sec:symm_match} shows that an existing practice is suboptimal: optimizing a symmetric divergence towards a KL-regularized target policy.
Appendix \ref{apdx:toy_examples} provides additional details and examples.

% Our discussion will be restricted to regularization Eq. ({\ref{eq:brpo_original}) only.
% where $\mu$ is a reference policy, $Q$ is the critic or target score used for improvement, $\tau > 0$ is a temperature, and $D$ is the divergence used for regularization.

\subsection{Symmetric Regularization Mitigates One-sided Bias}\label{sec:symm_reg}

\emph{Example 1: Regularization and One-sided Bias}.
This example shows that one-sided divergences used as regularizers can cause one-sided bias.
Consider a two-action bandit problem with behavior policy $\datasetPolicy = (0.98, 0.02)$ and $Q = (0,\Delta)$. 
We vary $\Delta \in [0,1]$ and plot the results in Figure \ref{fig:support-gap}(a)(b).
% \textbf{Example 1: Support Gap. }
% % \paragraph{Setting.}
% Consider a two-action bandit with behavior policy
% \[
% \mu = (0.98,\,0.02),
% \]
% true value function
% \[
% Q(a_1)=0,\qquad Q(a_2)=\Delta,
% \]
% and temperature $\tau = 0.5$ in \eqref{eq:regularized-objective}. The second action is optimal whenever $\Delta > 0$, but it is rare under the reference policy.
% This instance is intended to isolate the usual ``support gap'' objection to forward KL. If $\mu(a_2)$ were exactly zero, then any positive $\pi(a_2)$ would make $\mathrm{KL}(\pi\|\mu)$ infinite. I therefore keep $\mu(a_2)=0.02$ instead of zero so that the example stays finite while remaining very close to the hard support-mismatch regime.
% \begin{example}[Support gap]
% Consider a two-action bandit problem with behavior policy $\datasetPolicy = (0.98, 0.02)$ and $Q = (0,\Delta)$. 
% We sweep $\Delta \in [0,1]$.
The solution computed using the symmetric divergence $D_\mathrm{JS}(\pi||\datasetPolicy)$ consistently places more mass on the rare good action than the solution obtained from the asymmetric $D_\mathrm{KL}(\pi\|\datasetPolicy)$. 
For $\Delta=0.6$, for instance, we have
\[
\pi_{\mathrm{KL}}(a_2)=0.063, \,\, \mathbb{E}_{a \sim \pi_{\mathrm{KL}}}[Q(a)] = 0.038\qquad
\pi_{\mathrm{JS}}(a_2)=0.905, \,\, \mathbb{E}_{a \sim \pi_{\mathrm{JS}}}[Q(a)] = 0.543.
\]
% with the corresponding values
% \[
% \mathbb{E}_{a \sim \pi_{\mathrm{KL}}}[Q(a)] = 0.038,\qquad
% \mathbb{E}_{a \sim \pi_{\mathrm{JS}}}[Q(a)] = 0.543.
% \]
% \end{example}

The KL is one-sided: it heavily penalizes placing probability mass where the behavior policy is small (Figure \ref{fig:support-gap}(a)). 
In this case, this makes the improvement step too conservative (Figure \ref{fig:support-gap} (b)).
By contrast, JS does not punish the rare action solely through its ratio, showing that one-sided conservatism can be effectively mitigated.
To further support our claim, Appendix \ref{apdx:toy_examples} provides additional examples.
% \begin{interpretation}
% Forward KL is one-sided: it heavily penalizes placing probability mass where $\mu$ is tiny. In this instance that makes the improvement step too conservative, because the very direction we need to move is precisely the direction that forward KL suppresses. Jensen--Shannon is still anchored to $\mu$, but it does not punish the rare action solely through the ratio $\pi(a_2)/\mu(a_2)$. The example therefore isolates a clean setting in which symmetry helps because one-sided conservatism is itself the bottleneck.
% \end{interpretation}

\subsection{Symmetric Optimization Induces Better Boundary Policies}\label{sec:symm_opt}

\begin{figure}[t]
    \centering
    \includegraphics[width=\textwidth]{figs/toy_examples/support_boundary_compact.pdf}
    \caption{Symmetric divergences improve over one-sided divergences. (a)\&(b): symmetric regularization helps mitigate support gaps and bias. (c)\&(d): symmetric optimization improves policy behavior near the action boundary under clipping. 
    The percentage in (d) indicates the proportion of probability mass falling off $a=-1$.
    }
    \vspace{-1em}
    \label{fig:support-gap}
\end{figure}

\emph{Example 2: Boundary Fitting Under Action Clipping. }
In this example, we study the shape of a clipped policy near the action boundary.
This is relevant to many RL applications that clip actions outside a given range.
Neural network policies can have unexpected behavior when updated near the boundary due to clipping \citep{Lee2025-truncatedGaussianPolicy,Zhu2025-qExpPolicy}.
We construct an example with the action range clipped to $[-1,1]$ and a Gaussian $\pi_\theta$ as the behavior policy.
The goal is to learn the target policy, which is a mixture of truncated Gaussians concentrated near $a=-1$ (the black line in Figure \ref{fig:support-gap}(c)). 
The reward is also defined as a mixture of Gaussians.
See Appendix \ref{apdx:toy_examples} for full details.

%We set the action range to $[-1,1]$, which is common in many applications. We use a Gaussian policy $\pi_\theta$.
%The target policy is a mixture of truncated Gaussians concentrated near $a=-1$.
%Actions outside the boundary are clipped.
%The reward is also defined as a mixture of Gaussians.
%Appendix \ref{apdx:toy_examples} provides full details.

From Figure \ref{fig:support-gap}(c), we see that the asymmetric forward KL (red) is mass-covering and produces a wide distribution spilling a significant chunk of probability mass outside the allowed action range.
This results in a distorted actor policy because actions outside the range $a<-1$ will be clipped.
%Since actions outside $a<-1$ will be clipped, the actual actor policy will be distorted. 
In contrast, Jensen-Shannon penalizes mismatches on both sides, suppressing probability mass leakage outside the action range and thereby improving the post-clipping reward. Figure \ref{fig:support-gap}(d) shows that JS achieves roughly double the reward of KL.
% Martha: I don't understand what this sentence is saying. What percentage?
%The percentage indicates the proportion of probability mass falling outside $a=-1$.

\subsection{Matching the Symmetric Projection Geometry}\label{sec:symm_match}

% \textbf{Issue with Asymmetric $\regD$, Symmetric $\lossD$. }
Although there is an increasing amount of evidence that symmetric divergences used as $\lossD$ can improve agent performance \citep{Go2023-AligningLanguageModels-fDivergence,Li-2026-ChoiceDivergence-RLVROnline}, the
JS or Jeffreys divergences are often adopted to fit the KL-regularized target $\regPi_\text{KL}$.
These works assume that, because the function class $\pi_\theta$ is limited, a symmetric loss objective induces a different optimization landscape that could lead to better performance.
However, as we show below, this is generally suboptimal because a projection-geometry mismatch between symmetric $\lossD$ (e.g., JS or Jeffreys) and asymmetric $\regD$ (e.g., KL) can incur additional policy-improvement error.
\setcounter{theorem}{0}
\begin{theorem}[Matching a KL-regularized policy with JS]\label{thm:mismatch}
    Assume the target policy $\pi^*$ is KL-regularized. 
    Then minimizing a symmetric $\lossD(\pi^*||\pi_\theta)$ can incur additional policy representation error relative to minimizing a KL divergence loss.
\end{theorem}
\vspace{-1.5em}
\begin{proof}
    See Appendix \ref{apdx:thm_mismatch} for the proof. 
\end{proof}
\vspace{-0.5em}
 This is not merely a theoretical concern. Several RL algorithms have used a mismatched $\regD$ and
$\lossD$ \citep{Go2023-AligningLanguageModels-fDivergence,han-2025-fPO,Li-2026-ChoiceDivergence-RLVROnline}, and could therefore suffer from this issue.

\emph{Example 3: Different Solutions When $\regD \neq \lossD$. }
We consider a three-action bandit where the forward-KL-regularized target $\pi^*$ is matched using only two fixed $\pi_1, \pi_2$, by minimizing either the JS or the reverse KL divergence.
Figure \ref{fig:projection-geometry} illustrates the mismatched geometry induced by $\lossD$ and $\regD$.
Here, the black star represents the target $\pi^*$. 
Red contours show the geometry induced by $\regD$, and green contours show that induced by $\lossD$.
The first two subfigures show that under forward-KL regularization, mismatched $\lossD$  (JS and reverse KL) prefer $\pi_1$ (circled by green contours), yet the original regularization $\regD$ prefers $\pi_2$ (circled by brown contours). 
In the third subfigure, the brown curve is
a one-dimensional restricted policy curve, on which policies are only allowed to be projected. 
$\regPi$ is the matched projection, and
$\lossPi$ is the policy selected by the JS $\lossD$. 
In this case,  $\regD$ and $\lossD$ still prefer different solutions.
This example suggests that existing methods \citep{Go2023-AligningLanguageModels-fDivergence,Li-2026-ChoiceDivergence-RLVROnline} can be suboptimal when fitting a KL-regularized policy with JS.

\section{Symmetric Behavior Regularized Policy Optimization}\label{sec:issues}

Our goal is to establish the theory and practice of symmetric BRPO.
However, this is challenging with symmetric divergences because they offer neither a closed-form solution nor a numerically stable optimization objective.
We discuss these issues in Section \ref{sec:sbrpo-challenges}, present the details of our approach in Section \ref{sec:taylor}, and introduce a practical algorithm in Section \ref{sec:sfac}.

\begin{figure}[t]
    \centering
    \includegraphics[width=\textwidth]{figs/toy_examples/projection_geometry.pdf}
    \caption{
    A toy example of matching a forward-KL-regularized target $\pi^*$ with two candidates $\pi_1, \pi_2$, using either reverse KL or JS as $\lossD$.
    $\epsilon_\text{gap}$ denotes the gap between the solution of forward KL $\lossD$ and that of JS/reverse-KL.
    Red contours: geometry induced by forward KL regularization.
    Green contours: geometry of $\lossD$. 
With a limited policy class, mismatched $\regD \neq \lossD$ results in different solutions: $\pi_1$ is favored by $\lossD$, whereas $\pi_2$ is favored by $\regD$.
See Appendix \ref{apdx:mismatched_geometry} for detail.
}
\vspace{-1em}
    \label{fig:projection-geometry}
\end{figure}

\subsection{Challenges: No Analytic Policy and Numerical Instability}\label{sec:sbrpo-challenges}

We motivated symmetric divergences and the best practice of keeping $\regD = \lossD$ in Section \ref{sec:why_symm}. 
However, this becomes challenging for more complex tasks that require function approximation, as we show next.

\begin{theorem}\label{thm:no_analytic}
Assume we are given any symmetric $f$-divergence as defined in Definition \ref{def:wider_symm}.
% a continuously differentiable convex function $f: (0, \infty) \rightarrow [0, \infty]$ such that
% % $f(t)  = t\ln t + g(t)$,
% \begin{align*}
% f(t) & = t\ln t + g(t),
% \end{align*}
Then if $g'(t)$ does not make $f'(t)$ an affine function in $\ln t$, i.e. $f'(t) \neq a\ln t + b$,
the regularized optimal policy $\regPi$ does not have a closed-form expression.
\end{theorem}
%\vspace{-1.5em}
\begin{proof}
    See Appendix \ref{apdx:no_closed_form} for proof.
\end{proof}
It is worth noting that this affine form must be strictly in $\ln t$, excluding terms such as $\ln(t+1)$.
As an immediate corollary, the symmetric divergences in Table \ref{tab:f-divergences}, namely Jeffreys, Jensen-Shannon, and GAN, do not permit an analytic $\regPi$. This can be verified by checking that the Jeffreys divergence has $g'(t) = - \frac{1}{t}$, the Jensen-Shannon divergence has $g'(t) = -\ln \frac{t+1}{2} - 1$, and the GAN divergence has $g'(t) = -\ln(t+1) - 1$.
In Appendix \ref{apdx:review_no_closed} we show that existing characterizations such as DICE lead to the same conclusion.
Therefore, prior methods have focused exclusively on asymmetric candidates such as $\chi^2$ or KL as $\regD$ \citep{Wang2024-beyondReverseKL,han-2025-fPO,Li-2026-ChoiceDivergence-RLVROnline}.
Symmetric divergences used as $\lossD$ can also incur numerical issues, as we show in Theorem \ref{thm:issue_opt}, especially for (nearly) finite-support policies (i.e. $\exists\, a,\pi(a|s)=0$) that can be induced by Tsallis or $\alpha$-divergences. Such finite-support policies have become increasingly popular \citep{Li2023-quasiOptimal-qGaussian,Xu2023-OfflineNoODDAlphaDiv,Zhu2025-qExpPolicy}, making it important to address this issue with symmetric divergences.
% We next examine symmetric divergences as optimization objective Eq.(\ref{eq:brpo_optimize}).
% It turns out they can also be numerically problematic, especially for finite support policies (i.e. $\exists\, a,\pi(a|s)=0$) that can be induced by popular asymmetric divergences like the  Tsallis or $\alpha$-divergences \citep{Li2023-quasiOptimal-qGaussian,Xu2023-OfflineNoODDAlphaDiv,Zhu2025-qExpPolicy} that attract increasing interest in recent offline RL studies.
\begin{theorem}\label{thm:issue_opt}
    Minimizing symmetric divergence losses in Eq. (\ref{eq:brpo_optimize}) can incur numerical issues when $\regPi$ or $\pi_{\theta}$ are far apart or either has finite support.
\end{theorem}
\vspace{-1em}
\begin{proof}
    See Appendix \ref{apdx:numerical_issue} for proof.
\end{proof}
It is worth noting that even for full-support policies, the issue can persist when $\regPi \gg \pi_{\theta}$ and vice versa.
While some candidates like the Jensen-Shannon may not necessarily incur the same issue, existing research has reported instability \citep{Go2023-AligningLanguageModels-fDivergence}.
It is therefore important to find new tools to circumvent vanilla symmetric divergences.

\subsection{Symmetric BRPO via the Infinite Pearson-Vajda Series}\label{sec:taylor}
% \section{Tsallis KL Regularization as $D_{f_1}$}

In this section, we present the key theoretical contributions of the paper.
We propose a universal $f$-BRPO framework that addresses the lack of an analytic policy in Theorem \ref{thm:reg_solution}, a numerically stable surrogate for the vanilla symmetric optimization objective in Theorem \ref{thm:cond_symm}, and a bound on the difference between the surrogates and their counterparts in Theorem \ref{thm:error_bound}.
\begin{theorem}\label{thm:reg_solution}
All $f$-divergence BRPO objectives can be equivalently cast as follows:
\begin{align}
    % \regPi =  \argmax_{\pi} \,\, 
    \max_{\pi} \,\,
    \expectation{\substack{s\sim\mathcal{D}\\a\sim\pi}}{Q(s,a) - \tau  \sum_{n=0}^{\regN=\infty} \frac{f^{(n)}(1)}{n!}  \chiAny{n}{\policydot{\pi}}{\policydot{\datasetPolicy}}},
    \label{eq:taylor_reg}
\end{align}
where $\chi^n$ is the Pearson-Vajda $\chi$ divergence.
Further, if the series is truncated to $2 \le \regN < 5$, i.e. $n=0, 1, \dots, \regN$, then
% MArtha: these statements do not belong in a theorem statement. Put it outside of the statement
% This formulation can therefore replace direct symmetric regularization.
% However, by itself it does not yield an analytic policy $\pi^*$.
% Now let the series be truncated to $\regN<\infty$, i.e. $n=0, 1, \dots, \regN$, with $2\leq \regN<5$. 
% Then 
the regularized optimal policy $\regPi$ can be expressed analytically as
    % \begin{align*}
    %     \regPi (a|s) \propto \datasetPolicy(a|s) \AdaRectBracket{1 + Z_{\regN}^{\tau}(s,a)}_{+}  ,
    % \end{align*}
    % where $[\cdot]_{+} \!:=\! \max\{\cdot, 0\}$.
    % The solutions are:
    \begin{align}
    % \regPi (a|s) \propto \datasetPolicy(a|s) \,
    % \AdaRectBracket{ 1 + \frac{Q(s,a)}{\tau}}_{+}
    &\pi^*_{\regN=2}(a|s)  =  \mu(a|s) \AdaRectBracket{1 +  \frac{{Q(s,a)-\alpha(s)}}{2\tau}}_{+}, \nonumber\\
    &\pi^*_{\regN=3}(a|s) = \mu(a|s)\, \AdaRectBracket{
  1+\frac{-\tau_2 + \sqrt{\tau_2^{2} + 4\tau_3\,\AdaBracket{Q(s,a)-\alpha(s)}}}
       {2\tau_3} }_{+}, \label{eq:chi2_policy}\\
       & \pi^*_{\regN=4}(a|s) = \mu(a|s) \AdaRectBracket{
1+ \frac{1}{3\tau_4} \AdaBracket{ -\tau_3 + \sqrt[3]{\frac{R+\sqrt{R^2+4B^3}}{2}} + \sqrt[3]{\frac{R -\sqrt{R^2+4B^3}}{2}}}}_{+}.\nonumber
\vspace{-0.5em}
\end{align}
where $[\cdot]_{+} \!:=\! \max\{\cdot, 0\}$. $\alpha(s)$ is the normalization constant. 
$\tau_2 := 2\tau  \frac{f^{(2)}(1)}{2!}, \tau_3 := 3\tau \frac{f^{(3)}(1)}{3!}, \tau_4 := 4\tau \frac{f^{(4)}(1)}{4!}$, $B:=3\tau_4\tau_2-\tau_3^2,
R:=27\tau_4^2\AdaBracket{Q(s,a)-\alpha(s)}+9\tau_4\tau_3\tau_2-2\tau_3^3$.
% We refer the reader to Appendix \ref{apdx:reg_solution} for $\regN=4$.
% leading $\frac{f^{(2)}_\text{Reg}(1)}{2}$ to $\tau$.
\end{theorem}
\begin{proof}
    See Appendix \ref{apdx:reg_solution} for proof.
\end{proof}
Theorem \ref{thm:reg_solution} provides a tractable form of symmetric regularization when the series is finite.
With an infinite series, exact symmetric regularization is recovered, but no analytic policy exists.
As long as $\regN<5$, the Abel-Ruffini theorem \citep{Ramond-2022-AbelRuffni} guarantees the existence of an analytic solution. Note further that we must have $\regN \geq 2$ because $f^{(0)}(1)=0$ and $\chi^1 = 0$.

Let us reason more about choosing $\regN$ and $\tau$. 
The truncation $\AdaRectBracket{\cdot}_{+}$ suggests that the optimal policy is finite-support, with the truncation threshold controlled by $Z_N^{\tau}$, which depends on $\regN$ and $\tau$. 
We opt for $\regN=2$ for simplicity which still has the key truncation property.
% As shown by \citet{zhu2023generalized}, the threshold can in fact be fully controlled by $\tau$ alone.
% Therefore, it is safe to set $\regN=2$.
Table \ref{table:approx_table} lists the coefficients for the symmetric divergences.

For stable optimization,  we propose the following surrogate to replace vanilla symmetric $\lossD$.
% \begin{align}
%     \begin{split}
%         f(t) =    
%         % {f_\text{Asymm}(t)}  -   \,\,{f_\text{ConSym}(t)}\\
%          \underbrace{t\ln t}_\text{asymmetry}  +  \,\,\underbrace{g(t)}_{\substack{\text{conditional}\\\text{symmetry}}},
%     \end{split}
% \end{align}
% the first term $t\ln t$ corresponds to the asymmetric forward KL divergence.
% The second term $g(t)$, which can vary case by case, dictates how overall symmetry can be achieved given the first term.
% This decomposition allows us to improve numerical stability by expanding only $g(t)$, since $t\ln t$ leads to the numerically stable advantage regression, as have been shown in Eq.(\ref{eq:forward_kl_loss}).
% Therefore, we can decompose the optimization step as the following:

\begin{theorem}\label{thm:cond_symm}
    The following optimization objective approximates the symmetric divergences in Table \ref{tab:f-divergences} and avoids numerical issues provided that $\pi_\theta > 0$:
    \begin{align}
    \begin{split}
        \widehat{D}_\mathrm{Opt}(\theta) \!=\! 
        \expectation{(s,a)\sim\mathcal{D}\!}{-\AdaRectBracket{1 +\frac{Q(s,a) \!- \! V(s)}{\tau}}_{+} \ln \pi_{\theta}(a|s)} \!+ {\expectation{\substack{s\sim\mathcal{D}\\a\sim\pi_{\theta}\!}}{{ 
        \sum_{n=2}^{N_\text{loss}} \frac{f^{(n)}(1)}{n!} \! \AdaBracket{\frac{\regPi(a|s)}{\pi_{\theta}(a|s)} \!-\!1 }^n}}}.
    \end{split}
    \label{eq:taylor_loss}
\end{align}
\end{theorem}
\begin{proof}
    See Appendix \ref{apdx:conditional_symmetry} for proof.
\end{proof}

% Let us define the second term in $\lossD$ as $\mathcal{L}_{\text{ConSym}}(\theta)$, then
% By contrast, a full Taylor expansion involves purely powers of policy ratio which can be numerically problematic.
% here, the first common term $t\ln t$ corresponds to the asymmetric forward KL divergence.
% The second term, which varies from divergence to divergence, is the \emph{conditional symmetric} part that dictates how overall symmetry can be achieved given the first term.
% Given $\regPi$ as per Eq.(\ref{eq:chi2_policy}),
% we have seen in Eq.(\ref{eq:forward_kl_loss}) that the forward KL divergence can be cast as maximizing log-likelihood by sampling from the dataset, we only need to tackle the second term properly, i.e.

% Jensen-Shannon coefficient decays much faster than the Jeffrey's
% divergence could be numerically unstable as its coefficient decays at the rate of $n^{-1}$ which could be slow.

% Therefore, to minimize a symmetric loss divergence, it suffices to minimize the forward KL induced by $t \ln t$ and Eq.(\ref{eq:taylor_loss}) without needing to worry about support, since the actions are sampled from the learning policy $\pi_{\theta}$ itself.

It is worth noting that the threshold operator $[\cdot]_{+}$ filters out actions with value $Q(s,a) < V(s) -\tau$.
This stands in contrast to advantage weighted regression methods \citep{Peng2020-AdvantageWeighted} that assign a positive weight to any action. 
Thresholding low-valued actions has been shown to improve learning efficiency and performance \citep{Zhu2023-tsallisOffline}.
Eq. (\ref{eq:taylor_loss}) also has an interesting connection to a very recent method \citep{Huang2025-correctingMythos-Chi2} that proposed KL + $\chi^2$ regularization to improve RLHF alignment.
Despite the different settings, Eq. (\ref{eq:taylor_loss}) can be viewed as a generalization of their method, since we recover it by truncating the series at $N_\text{loss}=2$, with the coefficient $f^{(2)}(1)/2$ playing the role of the relative weighting.

Given the above surrogates, it is natural to ask how much they differ from the exact counterparts.
The following theorem provides a tight upper bound.
\begin{theorem}
\label{thm:error_bound}
Let $D^\infty$ denote the exact divergence and $\widehat{D}^N_\epsilon$ its $N$-term surrogate with ratio ${\regPi}/{\pi_{\theta}}$ clipped to the interval $[1-\epsilon, 1+\epsilon]$, on which $g^{(n)}$ is absolutely continuous. 
   Assume further the states are randomly sampled from the offline dataset, then
   % Let $\mathcal{L}^{\infty}$ denote the infinite series of $g(t)$ and $\widehat{\mathcal{L}}^{\regN}_{\epsilon}(\theta)$ the $\regN$-term truncated series with clipping, then 
    \begin{align*}
        \expectation{s\sim\mathcal{D}}{\left|{D}^{\infty}(\theta) - \widehat{D}^{N}_{\epsilon}(\theta) \right| }\leq \frac{2 \epsilon^{N+1}}{(N+1)!} \mynorm{g^{(N+1)}}{\infty} ,
    \end{align*}
    where $\mynorm{g^{(N+1)}}{\infty} := \sup_{t\in [1-\epsilon, 1+\epsilon]} \adaAbsolute{g^{(N+1)}(t)}$.
\end{theorem}
\begin{proof}
    See Appendix \ref{apdx:error_bound} for proof.
    \vspace{-0.75em}
\end{proof}
Theorem \ref{thm:error_bound} shows that the finite series is still a close approximation to the exact counterparts in regularization and optimization.
Consider the Jeffreys divergence ($g(t)=-\ln t$) with $\epsilon=0.2$ and  $N=5$; the upper bound becomes a very small number $8.13 \times 10^{-5}$.

\subsection{Symmetric $f$-Actor-Critic}\label{sec:sfac}

% The bound can be further improved if the dataset satisfies some regularity conditions.

To obtain a practical implementation, we assume for simplicity that at each policy update the action-value function $Q_{\psi}$ and the state-value function $V_{\phi}$  are available.
They are trained by standard TD learning.
We also need to be able to evaluate $\regPi(a|s)$ for actions sampled from $\pi_{\theta}$.
\begin{wrapfigure}[13]{r}{0.5\textwidth}
 % \vspace{-1em} 
\begin{algorithm}[H]
\caption{Symmetric $f$-Actor-Critic}\label{alg:sfac}
\KwIn{$s\sim\mathcal{D}$, $\tau>0, \regN=2, \lossN\geq 2$}
% Initialize policies by Alg. \ref{alg:init} \;
\While{\emph{learning}}{
    % sample  $(s, a)$ from dataset $\mathcal{D}$  \;
    % sample actions $a$ from behavior policy $\datasetPolicy$\;
    % compute $Q_{\psi}(s,a)$ and $V_{\phi}(s)$\;
    compute thresholded log-likelihood $\mathcal{L}_{t\ln t} :=$
    $- \widehat{\mathbb{E}}_{a\sim\mathcal{D}}\!\! \AdaRectBracket{ \AdaRectBracket{1+\frac{Q_{\psi}(s,a)-V_{\phi}(s)}{\tau}\!}_{+} \!\!\ln \pi_{\theta}(a|s) \!} $\;
    \vspace{-2mm}
    % sample $b$ from $\pi_{\theta}$ \;
    compute truncated series $\mathcal{L}_{g} :=$ $\widehat{\mathbb{E}}_{b\sim\pi_\theta}\!\! \AdaRectBracket{\sum_{n=2}^{N_\text{loss}} \!\frac{f^{(n)}(1)}{n!} \! \AdaBracket{\AdaRectBracket{\frac{\pi_{\zeta}(b|s)}{\pi_{\theta}(b|s)}}_{\epsilon} \!\!-\!1 }^n}$ \;
    update $\theta$ by minimizing $ \mathcal{L}_{t\ln t} + \mathcal{L}_{g}$\;
    update $\zeta$ by minimizing $\mathcal{L}_{t\ln t}$\;
  }
\end{algorithm}
  \vspace{-1em}                         % tighten spacing below
\end{wrapfigure}
To do this, we can either approximately evaluate $ \datasetPolicy(a|s)\AdaRectBracket{1+\frac{Q_{\psi}(s,a) - V_{\phi}(s)}{\tau}}_{+}$ without the normalization constant, which requires estimating $\datasetPolicy$, or parameterize $\regPi$ with another network $\zeta$ that is trained by advantage regression.
We find the latter approach to be more effective and stable in general.
Our algorithm Symmetric $f$-Actor-Critic (S$f$-AC) is described in Algorithm \ref{alg:sfac}, where $[\cdot]_{\epsilon} := \texttt{clip}(\cdot, 1-\epsilon, 1+\epsilon)$.

% Recall that we set the entropic index $q_w =0$ for the weighting coefficient $w(s,a)$, but in principle any $q_w < 1$ will have the filtering property.
% We initialize by Alg. \ref{alg:init} and sample from the policies by Alg. \ref{alg:sample}.

\section{Experiments}\label{sec:experiments}

We use the standard D4RL locomotion/Adroit/Maze benchmarks as the testbed for evaluating our method and performing ablation studies.
Appendix \ref{apdx:additional} provides additional  results.
Implementation details can be found in Appendix \ref{apdx:implementation}.

% However, we found that the same can happen for the S$f$-AC Jeffrey loss, which can also bias the distribution towards either mode of the target mixture depending on $\lossN$.
% By contrast, our JS tends to be stable across $\lossN$.

% Note that the results are in line with the lower bound property, that is, we
% have $D_{f} (P \rvert\lvert Q_{{\theta}^{\ast}}) \geq F(\hat{\omega}, \hat{\theta})$.
% There is a good correspondence between the gap in objectives and the difference
% between the fitted means and standard deviations. 
% The right side of Table~\ref{TableGMM} shows the results of the following
% experiment: (1) we train $T_{\omega}$ and $Q_{\theta}$ using a particular
% divergence, then (2) we estimate the divergence and re-train $T_{\omega}$ while
% keeping $Q_{\theta}$ fixed.
% As expected, $Q_{\theta}$ performs best on the divergence it was trained with.
% Further details showing detailed plots of the fitted Gaussians and the optimal
% variational functions are presented in the supplementary materials.

% In summary, the above results demonstrate that when the generative model is
% misspecified and does not contain the true distribution, the divergence
% function used for estimation has a strong influence on which model is learned.

\subsection{Standard Benchmarks}\label{sec:mujoco}

\begin{figure}[t]
    \centering
    \includegraphics[width=.875\textwidth]{figs/new/fac_baselines_d4rl.pdf}
    \vspace{-5pt}
    \caption{
    S$f$-AC Jensen-Shannon with $\lossN=3$ versus baselines on the D4RL MuJoCo environments.
    Solid lines show the mean, and shaded regions show the standard deviation, averaged over 5 seeds.
    Only S$f$-AC is shown with full opacity. 
    JS performs favorably compared to the baselines.
    }
    \vspace{-1em}
    \label{fig:d4rl}
\end{figure}

The D4RL MuJoCo suite has been a standard benchmark for testing various offline RL algorithms. 
In this section, we compare S$f$-AC Jensen-Shannon against the baselines on 9 environment-difficulty combinations. 
For S$f$-AC, we use $\lossN=3$ and perform a grid search over $\tau$ and learning rates; see Table \ref{table:offline_default_param}. 
For the baselines, we use their published settings.
All algorithms are run for $10^6$ steps and averaged over 5 seeds.

\begin{wrapfigure}[18]{r}{0.425\textwidth}
% \vspace{-1em}
\centering
    \includegraphics[width=.80\linewidth]{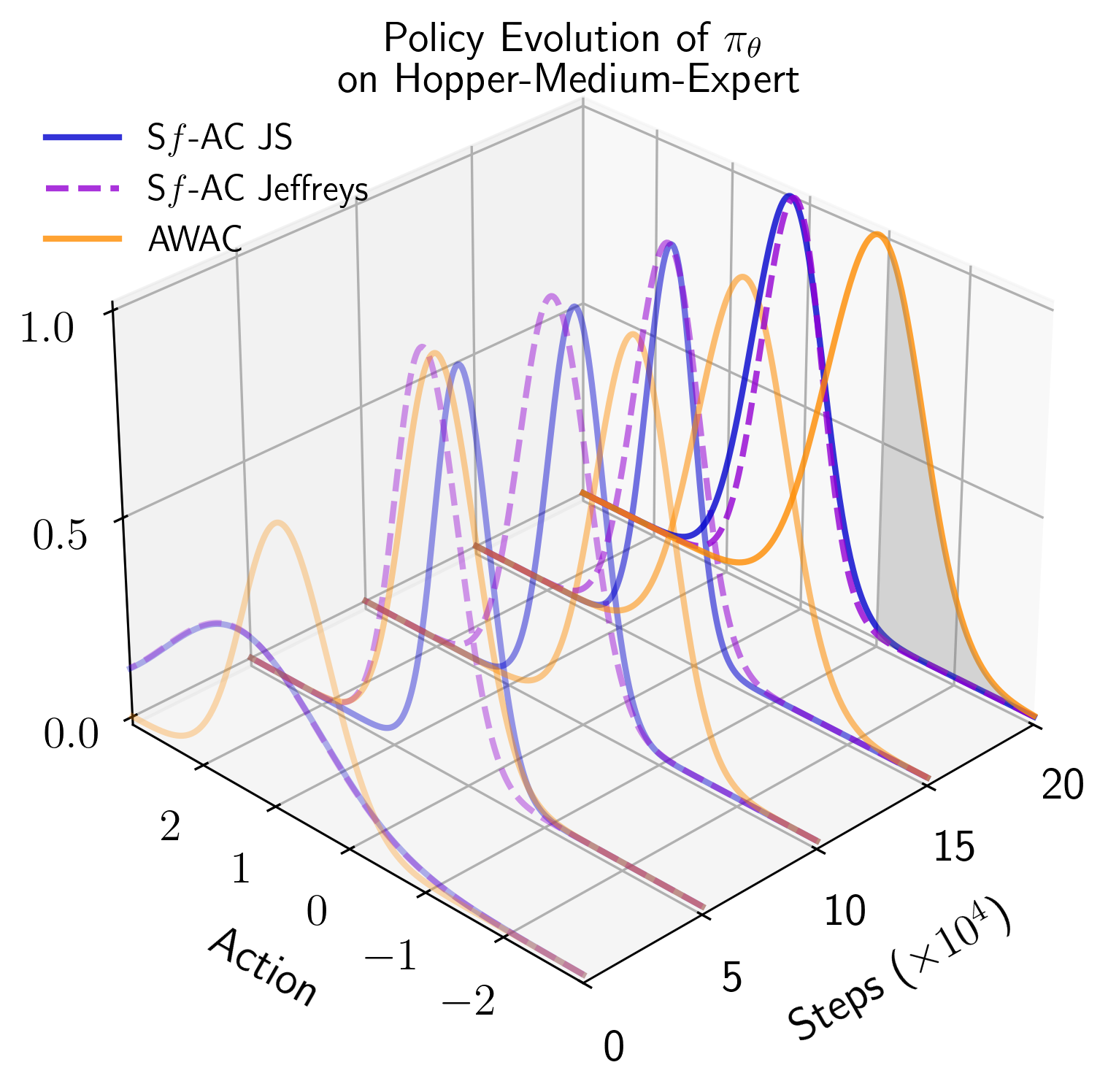}
    \vspace{-0.5em}
    \caption{
    Policy evolution of S$f$-AC versus forward KL minimization (AWAC). 
    AWAC increasingly prompts the policy beyond the minimum allowed action $-1$ (shaded area).
    }
    \label{fig:evolution}
\end{wrapfigure}

Figure \ref{fig:d4rl} shows the full results. 
Only S$f$-AC Jensen-Shannon is shown with full opacity.
The other methods are shown with transparency for a less cluttered visualization. 
Except on HalfCheetah-Medium and Hopper-Medium-Replay, S$f$-AC JS performs best or ranks among the top methods across all environments. 
Recall that the baselines are highly competitive on the MuJoCo locomotion tasks, and BPR \citep{Srinivasan-2025-BPR} requires extra resources and training time. Nevertheless, S$f$-AC performs favorably against these methods without incurring additional computational complexity; see Table \ref{tab:computation_time}.

% Table \ref{tab:d4rl} confirms this observation.
% Moreover, Figure \ref{fig:across_terms} demonstrates S$f$-AC is insensitive to the number of terms used to compute the symmetry divergence.
% This is desirable as S$f$-AC avoids the burdensome environment-specific parameter tuning to save resources and gain interpretability.

To further demonstrate the effectiveness of the proposed method, we compare S$f$-AC against popular baselines on the challenging Adroit and Maze2D benchmarks.
Table \ref{tab:maze2d-adroit} summarizes the result.
The reported numbers are the mean and standard deviation of the final scores, averaged over 5 seeds.
S$f$-AC JS is among the top performers in all cases.
Although S$f$-AC does not achieve the best score on maze2d-large, door-expert, and hammer-expert, its performance remains close to that of the best performer.

Figure \ref{fig:evolution} studies \emph{boundary policy fitting under action clipping} in a more challenging task.
We compare the policy evolution of S$f$-AC Jensen-Shannon (JS) and AWAC over the first 20\% of learning.
At step $0$, the policies are randomly initialized.
We include AWAC because it explicitly minimizes forward KL between $\regPi$ and $\pi_{\theta}$.
The allowed action range is $[-1, 1]$, but we show $[-2, 2]$ for better visualization.
AWAC spills a non-negligible chunk of mass outside $a<-1$, which can result in unintended policy behavior and affect performance \citep{Lee2025-truncatedGaussianPolicy}.  
By contrast, S$f$-AC keeps the majority of its probability mass within the allowed range, supporting the finding of Section \ref{sec:symm_opt} not only in tabular settings but also in more complex tasks.

% The former controls approximation to the symmetric divergence, and the latter controls the convergence of Taylor series.
% In this section we examine their effect in detail.
% Due to the page limit, we include additional results on generalized policy in Appendix \ref{apdx:additional}.

% It is visible that both JS and Jeffreys keep policies in the range, while the optimizing forward KL of AWAC increasingly prompts the policy beyond the allowed minimum action $-1$ (shaded area).
% Since actions outside the range are clipped (in this case $-1$), the policy will have an unintended shape and cause significant bias.
% % AWAC increasingly renders the policy a bang-bang policy \citep{Seyde2021-BangBangController}

% Preamble:
% \usepackage{booktabs}
% \usepackage[table]{xcolor}

% Preamble:
% \usepackage{booktabs}
% \usepackage[table]{xcolor}
% \definecolor{bestblue}{RGB}{221,235,247}

% Preamble:
% \usepackage{booktabs}
% \usepackage[table]{xcolor}
% \definecolor{bestblue}{RGB}{221,235,247}

% Preamble:
% \usepackage{booktabs}
% \usepackage[table]{xcolor}
% \definecolor{bestblue}{RGB}{221,235,247}

\begin{table*}[t]
\centering
\scriptsize
\setlength{\tabcolsep}{3.5pt}
\resizebox{\textwidth}{!}{%
\begin{tabular}{lccccccc}
\toprule
Task & S$f$-AC JS & TD3+BC & CQL & IQL & ReBRAC & EDAC & DT \\
\midrule
maze2d-umaze &
99.96 $\pm$ 22.41 &
29.41 $\pm$ 12.31 &
-8.90 $\pm$ 6.11 &
42.11 $\pm$ 0.58 &
\cellcolor{bestblue}\textbf{106.87 $\pm$ 22.16} &
95.26 $\pm$ 6.39 &
18.08 $\pm$ 25.42 \\
maze2d-medium &
\cellcolor{bestblue}\textbf{107.80 $\pm$ 36.50} &
59.45 $\pm$ 36.25 &
86.11 $\pm$ 9.68 &
34.85 $\pm$ 2.72 &
105.11 $\pm$ 31.67 &
57.04 $\pm$ 3.45 &
31.71 $\pm$ 26.33 \\
maze2d-large &
92.07 $\pm$ 82.87 &
\cellcolor{bestblue}\textbf{97.10 $\pm$ 25.41} &
23.75 $\pm$ 36.70 &
61.72 $\pm$ 3.50 &
78.33 $\pm$ 61.77 &
95.60 $\pm$ 22.92 &
35.66 $\pm$ 28.20 \\
\midrule
door-expert &
106.46 $\pm$ 36.95 &
-0.33 $\pm$ 0.01 &
-0.32 $\pm$ 0.02 &
\cellcolor{bestblue}\textbf{106.65 $\pm$ 0.25} &
106.37 $\pm$ 0.29 &
106.29 $\pm$ 1.73 &
104.87 $\pm$ 0.39 \\
pen-expert &
\cellcolor{bestblue}\textbf{173.56 $\pm$ 26.08} &
122.53 $\pm$ 21.27 &
-1.41 $\pm$ 2.34 &
128.05 $\pm$ 9.21 &
152.16 $\pm$ 6.33 &
-1.55 $\pm$ 0.81 &
116.38 $\pm$ 1.27 \\
relocate-expert &
\cellcolor{bestblue}\textbf{113.32 $\pm$ 4.65} &
-1.73 $\pm$ 0.96 &
-0.30 $\pm$ 0.02 &
106.11 $\pm$ 4.02 &
107.52 $\pm$ 2.28 &
71.94 $\pm$ 18.37 &
104.28 $\pm$ 0.42 \\
hammer-expert &
129.09 $\pm$ 2.45 &
3.11 $\pm$ 0.03 &
0.26 $\pm$ 0.01 &
128.68 $\pm$ 0.33 &
\cellcolor{bestblue}\textbf{133.62 $\pm$ 0.27} &
28.52 $\pm$ 49.00 &
117.45 $\pm$ 6.65 \\
\bottomrule
\end{tabular}%
}
\caption{Comparison between S$f$-AC Jensen-Shannon against popular baselines on Adroit and Maze2D benchmarks.
Shown numbers are last scores mean and std, averaged over 5 seeds.
The best scores are highlighted.
The proposed method performs the best or is among the top performers. 
}
\vspace{-1em}
\label{tab:maze2d-adroit}
\end{table*}

% Preamble:
% \usepackage{booktabs}
% \usepackage[table]{xcolor}
% \definecolor{bestblue}{RGB}{221,235,247}

% \begin{table}[t]
% \centering
% \small
% \setlength{\tabcolsep}{5pt}
% \begin{tabular}{llcl}
% \toprule
% Task & Best FAC HP $(\alpha,\eta_Q)$ & FAC $\max(\texttt{last\_score})$ & CORL best score \\
% \midrule
% maze2d-medium-v1   & ($10^{-4}$, $10^{-4}$)        & 136.18 & \cellcolor{bestblue}\textbf{EDAC: $154.41 \pm 1.58$} \\
% maze2d-umaze-v1    & ($10^{-3}$, $3\times10^{-4}$) & 135.61 & \cellcolor{bestblue}\textbf{ReBRAC: $162.28 \pm 1.79$} \\
% maze2d-large-v1    & ($10^{-4}$, $3\times10^{-4}$) & \cellcolor{bestblue}\textbf{254.52} & AWAC: $227.31 \pm 1.47$ \\
% \midrule
% door-expert-v1     & ($10^{-1}$, $3\times10^{-3}$) & 106.46 & \cellcolor{bestblue}\textbf{EDAC: $109.22 \pm 0.24$} \\
% pen-expert-v1      & ($10^{-1}$, $3\times10^{-4}$) & \cellcolor{bestblue}\textbf{173.56} & AWAC: $162.53 \pm 0.30$ \\
% relocate-expert-v1 & ($10^{-1}$, $10^{-4}$)        & \cellcolor{bestblue}\textbf{113.32} & AWAC: $111.21 \pm 0.32$ \\
% hammer-expert-v1   & ($10^{-1}$, $10^{-4}$)        & 129.09 & \cellcolor{bestblue}\textbf{ReBRAC: $134.74 \pm 0.30$} \\
% \bottomrule
% \end{tabular}
% \caption{FAC against CORL best scores on Maze2D and Adroit. For each task, the FAC row is chosen by maximizing $\max(\texttt{last\_score})$ over $(\alpha,\eta_Q)$, where $\eta_Q$ is the critic learning rate. CORL values are the strongest reported entries from the \emph{Best Scores} section of the CORL offline benchmark.}
% \label{tab:fac-best-adroit-maze}
% \end{table}

\subsection{Ablation Studies}\label{sec:ablation}

Since we can fix $\regN = 2$ by Eq.(\ref{eq:chi2_policy}), the remaining hyperparameter introduced by our algorithm is $\lossN$ in Eq. (\ref{eq:taylor_loss}).
$\lossN$ decides the number of terms that are used to expand the conditional symmetry term. 
We sweep S$f$-AC Jensen-Shannon (JS) and Jeffreys over different values of $\lossN$ from 2 to 6 on six environments and show the resulting trends in Figure \ref{fig:ablation}.
Curves show the means and std of last scores along the term-count axis.
The shaded regions indicate the environment-specific best $\lossN$.
Some environments, such as Relocate-Expert, prefer a larger number of terms, suggesting the dataset could be challenging: the target and the actor could differ significantly and hence higher-order terms outgrows the coefficients in Table \ref{table:approx_table},  contributing non-negligibly to the symmetric-divergence approximation.
Overall, reasonable performance can be expected by choosing $\lossN \in \{2, 6\}$ across environments, as our algorithm is moderately robust to the value of this hyperparameter.

\begin{figure}[t]
    \vspace{-0.5em}
    \centering
    \includegraphics[width=0.85\linewidth]{figs/new/option_b_trend.pdf}
    \vspace{-0.5em}
    \caption{
     The effect of the number of terms $\lossN$ on S$f$-AC Jensen-Shannon, shown for six environments.
     Curves show the mean and std of last scores. 
     The shaded region indicates the environment-specific best $\lossN$. 
    }
    \label{fig:ablation}
    % \vspace{-1em}
\end{figure}

% \vspace{-10pt}

% \textbf{Generalized parametric policies. }
% In Theorem \ref{thm:reg_solution} we derived the solution of $\regPi$ which is a finite-support distribution that can yield zero probabilities for some actions.
% In RL, this type of policies has been shown to connect to the general entropy regularization \citep{Lee2018-TsallisRAL,Nachum18a-tsallis,Li2023-quasiOptimal-qGaussian}.
% Specifically, \citet{zhu2023generalized,Zhu2025-FatToThin} showed that the $q$-exp functions for various $q<1$ can achieve the same sparsity effect by controlling the regularization strength $\tau$.
% In this paper we focused on $N=2$ which corresponds to the standard case $q=0$.
% \citet{Nielsen2013-chiApproxFdiv} showed that 5 to 7 terms are sufficient for close approximation.

\section{Related Work}\label{sec:related_work}

The functional form of $\regPi$ may vary depending on the regularizer.
The policy  $\regPi$ might be difficult to compute exactly and sample from.
Therefore, it is typically used as the distribution match target via $\lossD$. 
The existing literature  \citep{Ma2022-offlineGCRL-fAdvantage,Agarwal2023-f-PolicyGradients,zhu2023generalized}  has mostly focused on asymmetric divergences such as the KL, $\chi^2$, Tsallis divergences that permit a closed-form regularized solution.
Their implementations are straightforward, since the objective can be derived by computing only $f'$ and the score function. 
Recently, it has been increasingly noted that employing a symmetric divergence as $\lossD$ can improve agent's performance \citep{Go2023-AligningLanguageModels-fDivergence,Wang2024-beyondReverseKL,han-2025-fPO,Li-2026-ChoiceDivergence-RLVROnline}.
However, their target policy is still KL-regularized.
As we have demonstrated in Theorem \ref{thm:mismatch}, this can be suboptimal than matching a regularized policy under the same symmetric divergence regularizer.

Our method leverages the infinite series representation of Pearson-Vajda divergences. 
Some methods also adopted a similar idea of infinite series but to serve completely purposes.
\citet{Tang2020-TaylorExpansionPolicyOptimization} expanded the action-value difference in the transition dynamics and drew an analogy to the series.
\citet{Omura2024-TaylorExpansionStabilizeExtremeQLearning} proposed to expand the exponential function in the extreme Q-learning objective \citep{Garg2023-extremeQlearning} into a MacLaurin series to stabilize learning.
% By contrast, we utilized the series to obtain an analytic policy expression and a stable optimization surrogate.
% In this paper, we utilize the Taylor expansion of an $f$-divergence to obtain a $\chi^n$ series, and based on it derive an analytic policy $\regPi$ and a tractable minimization objective.

% In other areas such as the goal-conditioned RL \citep{Ma2022-offlineGCRL-fAdvantage,Agarwal2023-f-PolicyGradients} or RLHF \citep{Go2023-AligningLanguageModels-fDivergence,Wang2024-beyondReverseKL}, symmetric $f$-divergences have been discussed since they require only minimizing the divergence as a loss objective and no $\regPi$ is required.
% Their objective can be derived by computing only $f'$ and gradient of log-likelihood.
% $f$-Advantage regression \citep{Agarwal2023-f-PolicyGradients,Go2023-AligningLanguageModels-fDivergence}, no closed-form policy needed in LLM \citep{Wang2024-beyondReverseKL,Huang2025-correctingMythos-Chi2}

\section{Conclusion}

% Behavior Regularized Policy Optimization (BRPO) leverages asymmetric divergence regularization to mitigate  distribution shift in offline Reinforcement Learning. 
This paper is the first to study the open question of symmetric BRPO. 
We showed that symmetric divergences can outperform asymmetric ones such as KL in handling one-sided bias, near-boundary policy updates, and projection geometry consistency.  
We also showed that the existing practice of matching a KL-regularized target with JS can be suboptimal.
To develop the theory and practice of symmetric BRPO, several challenges needed to be addressed:
symmetric divergences do not permit a closed-form solution when used as regularizers,
and can incur numerical instabilities when used as optimization objectives.
We proposed a novel universal BRPO framework based on the infinite Pearson-Vajda divergence series and used it to represent symmetric BRPO.
With a finite-series representation, we obtained
(1) a closed-form optimal policy expression;
(2) a numerically stable optimization surrogate; and 
(3) a tight bound on approximation quality.
We compared the proposed method against several offline algorithms on the D4RL benchmark and observed consistently strong results, opening up possibilities for more diverse and effective divergence choices for BRPO. A limitation of our work is that the infinite series of Pearson-Vajda divergences needs to stay close to the convergence radius $1$, which we enforced by PPO-style clipping. 
We leave it to future work to find a way to avoid the computation of the policy ratio, and therefore the clipping.

% \section*{References}

% \newpage

\bibliographystyle{abbrvnat}
\bibliography{library}

\newpage
%%%%%%%%%%%%%%%%%%%%%%%%%%%%%%%%%%%%%%%%%

\appendix

\part*{Appendix}
\addcontentsline{toc}{part}{Appendix}
\pdfbookmark[1]{Appendix}{appendix}

\etocsettocstyle{\section*{}}{}
\etocsetnexttocdepth{2} % sections + subsections
\localtableofcontents
% \addcontentsline{toc}{section}{Appendix} % Add the appendix text to the document TOC
% \part{Appendix} % Start the appendix part
% \parttoc % Insert the appendix TOC
\newpage

\section{Mathematical Details and Proofs}\label{apdx:math_details}

\subsection{Proof of Theorem \ref{thm:mismatch}}\label{apdx:thm_mismatch}

To prove Theorem \ref{thm:mismatch}, we first prove the following more general version and recover Theorem \ref{thm:mismatch} as a special case, namely that KL $\regD$ with symmetric $\lossD$ is suboptimal.
For ease of notation, we use vector representation, so that $\langle p, q\rangle = \expectation{p}{q}$.
\setcounter{theorem}{0}
\begin{theorem}[Matching Projection Geometry under Limited Policy Class]
Consider the regularized policy improvement objective
\[
\mathcal{J}_{\mathrm{Reg}}(\pi)
=
\mathbb \langle Q, \pi\rangle -\tau D_{\mathrm{Reg}}(\pi\|\pi_{\mathcal D}).
\]
Let $\pi^*_{\mathrm{Reg}}$ be a maximizer of $\mathcal{J}_{\mathrm{Reg}}$. 
Given a limited policy class $\Pi$ (e.g. Gaussian), we can define the objective gap as:
\begin{equation}
G_{\mathrm{Reg}}(\pi)
:=
\mathcal{J}_{\mathrm{Reg}}(\pi^*_{\mathrm{Reg}})
-
\mathcal{J}_{\mathrm{Reg}}(\pi)
\ge 0.
\label{eq:general-gap}
\end{equation}
Then the best solution within the class is $\pi^*_{\Pi,\mathrm{Reg}}
\in \arg\min_{\pi\in\Pi}G_{\mathrm{Reg}}(\pi)$.
If an alternative optimization objective $D_{\mathrm{Opt}}$ is used, its solution is $\hat\pi_{\mathrm{Opt}}
\in
\arg\min_{\pi\in\Pi}D_{\mathrm{Opt}}(\pi;\pi^*_{\mathrm{Reg}})$.
We can therefore define the gap between them as:
\begin{equation}
\varepsilon_{\mathrm{gap}}
=
G_{\mathrm{Reg}}(\hat\pi_{\mathrm{Opt}})
-
\min_{\pi\in\Pi}G_{\mathrm{Reg}}(\pi)
\ge 0,
\label{eq:gap-error}
\end{equation}
with equality if and only if $\hat\pi_{\mathrm{Opt}}$ is also a minimizer of $G_{\mathrm{Reg}}$ over $\Pi$.
By further assuming that $D_{\mathrm{Reg}}(\pi\|\pi_{\mathcal D})$ is a Bregman divergence
\[
D_{\Omega}(\pi || \datasetPolicy)
=
\Omega(\pi)-\Omega(\datasetPolicy)-\langle\nabla\Omega(\datasetPolicy),\pi-\datasetPolicy\rangle,
\]
where $\Omega$ is a continuously differentiable, strictly convex function.
Then the gap is:
\begin{align}
G_{\mathrm{Reg}}(\pi) &= \tau D_{\Omega}(\pi\|\pi^*_{\mathrm{Reg}}). \nonumber \\
\Rightarrow \, \, \text{if} \quad \pi^*_{\Pi,\mathrm{Reg}}
&\in \arg\min_{\pi\in\Pi}G_{\mathrm{Reg}}(\pi) \nonumber \\
\text{then} \quad \pi^*_{\Pi,\mathrm{Reg}} &\in
\arg\min_{\pi\in\Pi}D_{\Omega}(\pi\|\pi^*_{\mathrm{Reg}}).
\label{eq:bregman-gap}
\end{align}
Therefore, minimizing the gap is equivalent to minimizing the corresponding Bregman divergence:
\begin{equation}
\pi^*_{\Pi,\mathrm{Reg}}
\in
\arg\min_{\pi\in\Pi}D_{\Omega}(\pi\|\pi^*_{\mathrm{Reg}}).
\label{eq:exact-bregman-projection}
\end{equation}
As a result, if $\hat{\pi}_\mathrm{Opt}$ does not minimize the gap, it incurs extra error. 
Corollary \ref{cor:klreg_jsopt} shows a concrete example.

\end{theorem}

\begin{proof}
$\mathcal{J}_{\mathrm{Reg}}(\pi^*_{\mathrm{Reg}})$ is constant over $\Pi$, so maximizing $\mathcal{J}_{\mathrm{Reg}}(\pi)$ over $\Pi$ is equivalent to minimizing the gap in Eq. (\ref{eq:general-gap}).
To show the Bregman part, notice that for the objective
\[
\mathcal{J}_{\Omega}(\pi^*_{\mathrm{Reg}}) = \mathbb \langle \pi^*_{\mathrm{Reg}}, Q \rangle -\tau D_{\Omega}(\pi^*_{\mathrm{Reg}}\|\pi_{\mathcal D}),
\]
the first-order optimality yields
\[
Q^*=\tau\big(\nabla\Omega(\pi^*_{\mathrm{Reg}})-\nabla\Omega(\pi_{\mathcal D})\big)+\lambda\mathbf 1
\]
for some scalar $\lambda$, where $\mathbf{1}$ denotes an all-one vector.  Hence
\begin{align}
\mathcal{J}_{\Omega}(\pi^*_{\mathrm{Reg}})
- \mathcal{J}_{\Omega}(\pi) &=
\langle Q,\pi^*_{\mathrm{Reg}}-\pi\rangle -\tau\big(
D_{\Omega}(\pi^*_{\mathrm{Reg}}\|\pi_{\mathcal D})
-D_{\Omega}(\pi\|\pi_{\mathcal D}) \big)\notag\\ 
&=\tau D_{\Omega}(\pi\|\pi^*_{\mathrm{Reg}}),
\end{align}
where the last step is the three-point identity for Bregman divergences together with the simplex constraint $\langle \mathbf 1,\pi^*_{\mathrm{Reg}}-\pi\rangle=0$, which concludes the proof.
\end{proof}

\begin{corollary}[Minimizing JS towards KL-regularized target]\label{cor:klreg_jsopt}
Let us consider the KL-regularized objective:
\[
\mathcal{J}_{\mathrm{Reg}}(\pi) = \mathbb \langle \pi, Q \rangle -\tau D_\mathrm{KL}(\pi\|\pi_{\mathcal D}),
\]
then the \emph{matched} optimization objective is
\[
\pi^*_{\Pi,\mathrm{Reg}} \in \arg\min_{\pi\in\Pi}D_\mathrm{KL}(\pi\|\pi^*_{\mathrm{Reg}}).
\]
If instead JS divergence is used, i.e. $\hat\pi_{\mathrm{JS}} \in \arg\min_{\pi\in\Pi}D_\mathrm{JS}(\pi \|\pi^*_{\mathrm{Reg}})$
then it is generally solving a different projection problem, and it incurs extra policy improvement error:
\[
\varepsilon_{\mathrm{gap}}
=
\mathrm{KL}(\hat\pi_{\mathrm{JS}}\|\pi^*_{\mathrm{Reg}})
-
\min_{\pi\in\Pi}\mathrm{KL}(\pi\|\pi^*_{\mathrm{Reg}})
\ge 0.
\]
The inequality is strict if the JS solution is not also a minimizer of $D_\mathrm{KL}(\pi\|\pi^*_{\mathrm{Reg}})$ over $\Pi$.
\end{corollary}

\begin{proof}
Take $\Omega(p)=\langle p, \log p \rangle$, the induced Bregman divergence is $D_\mathrm{KL}(p\|q)$. 
Then Eq. (\ref{eq:exact-bregman-projection}) becomes the matched forward-KL fit to $\pi^*_{\mathrm{Reg}}$. 
The displayed expression for $\varepsilon_{\mathrm{gap}}$ follows from \eqref{eq:gap-error} with $D_{\mathrm{Opt}}(\pi ||\pi^*_{\mathrm{Reg}})=D_\mathrm{JS}(\pi ||\pi^*_{\mathrm{Reg}})$. Strict positivity holds whenever the JS minimizer is not a KL minimizer.
\end{proof}

\subsection{Proof of Theorem \ref{thm:no_analytic}}\label{apdx:no_closed_form}

Let us write out the Lagrangian of the objective \citep{Li2019-regularizedSparse,Xu2023-OfflineNoODDAlphaDiv}:
\begin{align}
\mathcal{L}(\pi, \alpha, \beta) = 
&\sum_s d^{\datasetPolicy}(s)\sum_{a}\AdaRectBracket{\pi(a|s)
Q(s,a)
  -
  \tau\, 
  % \AdaBracket{ \frac{\pi(a|s)}{\mu(a|s)}  - 1}^N
  \expectation{\mu}{f\AdaBracket{\frac{\pi(a|s)}{\mu(a|s)}}}
}  \nonumber \\
&
-\sum_s d^{\datasetPolicy}(s)
\!\AdaRectBracket{\alpha(s)\AdaBracket{\sum_{a}\pi(a| s)-1}
-\sum_{a}\beta(a|s)\,\pi(a|s)}.
\label{eq:f_lagrangian}
\end{align}
where $d^{\datasetPolicy}$ is the stationary state distribution of the behavior policy.
$\alpha$ and $\beta$ are Lagrangian multipliers for
the equality and inequality constraints, respectively.
The KKT conditions are:
\begin{align*}
&\text{Primal feasibility:}
\quad  \sum_{a}\pi(a| s)=1,
\quad \pi(a| s)\ge0, \\[3pt]
&\text{Dual feasibility:}
\quad  \beta(a|s)\ge 0 , \\[3pt]
&\text{Stationarity:}
\quad 
\frac{\partial\mathcal{L}}{\partial \pi(a| s)}
= Q(s,a)
-\tau\, \mu(a|s)f'\AdaBracket{\frac{\pi(a|s)}{\mu(a|s)}} \frac{1}{\mu(a|s)} -\alpha(s) + \beta(a|s)=0,\\[3pt]
&\text{Complementarity:}
\quad  \beta(a|s)\,\pi(a| s)=0 
\end{align*}
Following \citep{Li2019-regularizedSparse,Xu2023-OfflineNoODDAlphaDiv}, we eliminate $d^{\datasetPolicy}$ since we assume all policies induce an irreducible Markov chain.
For any action with $\pi(a|s)>0$, we have $\beta(a|s)=0$.
Therefore, we can derive the solution as:
\begin{align*}
    &f'\AdaBracket{\frac{\pi(a|s)}{\mu(a|s)}} = \frac{Q(s,a) - \alpha(s)}{\tau} \quad \Rightarrow \quad \pi^*(a|s) = \AdaRectBracket{(f')^{-1}\!\AdaBracket{\frac{Q(s,a) - \alpha(s)}{\tau}} }_{+},
\end{align*}
where $\alpha(s)$ is the normalization constant that ensures $\pi^*$ sums to 1.

For $\pi^*$ to be analytic, we need to know how to compute $\alpha(s)$.
Since $f(t)$ begins with a $t \ln t$ term, so $f'(t) = \ln t + 1 + g'(t)$. If $g'(t)$ is not a function such that $f'(t) = a\ln t + b$ for some constants $a,b$, then $\alpha(s)$ cannot be calculated from the constraint $\sum_{a}\pi^*(a|s)=1$.

\subsection{Proof of Theorem \ref{thm:issue_opt}}\label{apdx:numerical_issue}

This section provides examples showing minimizing vanilla JS and Jeffreys can incur numerical issues.
For simplicity, we assume the following one dimensional case:
\begin{align*}
    p=\mathcal N(0,1), \qquad q_\theta=\mathcal N(\theta,1),
\end{align*}
i.e. $p$ is the target. 
The actor $q$ is parametrized as a Gaussian:
where $\theta\in\mathbb R$ is the parameter.
As $|\theta|$ grows, the divergence approaches its maximum $\log 2 \approx 0.6931$.
By varying $\theta$, we have
\begin{align*}
&\theta=4: \quad D_\mathrm{JS}(p\|q_\theta)  \approx 0.6327, \quad \frac{d}{d\theta}D_\mathrm{JS}(p\|q_\theta)\approx 6.86\times 10^{-2},
\\
&\theta=6: \quad D_\mathrm{JS}(p\|q_\theta) \approx 0.6893, \quad
\frac{d}{d\theta}D_\mathrm{JS}(p\|q_\theta)\approx 6.23\times 10^{-3},
\\
& \theta=8: \quad D_\mathrm{JS}(p\|q_\theta) \approx 0.6931, \quad
\frac{d}{d\theta}D_\mathrm{JS}(p\|q_\theta)\approx 1.96\times 10^{-4}.
\end{align*}
Therefore, we can see that the JS loss approaches the maximum possible value at $\theta=8$, but the gradient with respect to the Gaussian mean is nearly zero.
Vanilla JS as an optimization objective can be problematic when the target policy and the actor are far apart, which is often the case at the beginning of learning.
The following example shows how underflow can happen under JS.
Let
\begin{align*}
    p=\mathcal N(0,0.1^2), \qquad q=\mathcal N(8,0.1^2).
\end{align*}
Then at the middle point $a=4$,
\begin{align*}
    \log p(4) &= \log q(4)=-798.616.\\
    &\Rightarrow \exp  (-798.616) \approx 0 . \qquad \text{(underflow)}
\end{align*}
Therefore, the mean of the two policies is $m(a) = \frac{1}{2}(p(a) + q(a)) = 0$, producing $\log m(a) = -\infty$, leading to undefined JS value.

For  Jeffreys, it is clear from the definition that whenever $p$ or $q$ has finite support, the divergence is undefined. 
Even when both policies have full support, Jeffreys can result in overly aggressive updates.
To see this, assume the actor has log-std parametrization $\theta = \log \sigma$ which is standard.
Then 
\begin{align*}
    p=\mathcal N(0,1), \qquad q_\theta=\mathcal N(0,e^{2\theta}),
\end{align*}
Computing Jeffreys is equivalent to computing forward and reverse KL for two Gaussians, giving
Numerically,
\begin{align*}
    &\theta=3: \quad J \approx 200, \qquad \frac{d}{d\theta}J \approx 403.4,\\
    & \theta=5: \quad J \approx 11012, \qquad \frac{d}{d\theta}J  \approx 22026.5.
\end{align*}
I.e., Jeffreys can become extremely steep in the variance direction even when both policies have full support, which can in turn lead to aggressive updates and get stuck in local optima.

\subsection{Proof of Theorem \ref{thm:reg_solution}}\label{apdx:reg_solution}

% \textbf{Theorem \ref{thm:reg_solution}.}
% \emph{
%     Let the series in Eq.(\ref{eq:taylor_reg}) be truncated to $2\leq N<5$. 
%     Then the regularized optimal policy $\regPi$ can be expressed analytically.
%     Moreover, when $N=2$, the solution is 
%     \begin{align}
%     \regPi (a|s) \propto \datasetPolicy(a|s) \AdaRectBracket{\frac{1}{2} + \frac{Q(s,a) }{2\tau}}_{+} \propto\, \datasetPolicy(a|s) \,\exp_{q=0}{\AdaBracket{\frac{Q(s,a)}{\tau}}} ,
%     \nonumber
%     % \label{eq:chi2_policy}
% \end{align}
% where we absorbed the leading $\frac{f^{(2)}_\text{Reg}(1)}{2}$ to $\tau$.
% }
To prove Theorem \ref{thm:reg_solution}, we need to first show that any $f$-divergence regularization can be equivalently cast as regularization with an infinite series in $\chi^n$ divergence
.
The first part comes directly from following classic result establishing equivalence between a $f$-divergence and its Taylor series.
\begin{lemma}[\citep{Nielsen2013-chiApproxFdiv}]\label{lemma:taylor}
Let $f$ satisfies the conditions in the main text, further let $f$ be $n$-times differentiable.
Then a valid $f$-divergence permits the following Taylor expansion
    \begin{align*}
    D_{f}\AdaBracket{\policydot{\pi} || \policydot{\datasetPolicy}} = \int \datasetPolicy(a|s) \sum_{n=0}^{\infty} \frac{f^{(n)}(1)}{n!}  \AdaBracket{\frac{{\pi(a|s)}}{{\datasetPolicy(a|s)}} - 1}^n \mathrm{d}a,
        % D_{f}\AdaBracket{\policydot{\pi} || \policydot{\datasetPolicy}} = \int \datasetPolicy(a|s) \sum_{n=0}^{\infty} \frac{f^{(n)}(1)}{n!}      \chiAny{n}{\policydot{\pi}}{\policydot{\datasetPolicy}} \mathrm{d}a,
\end{align*}
where $f^{(n)}$ denotes the $n$-th order derivative of $f$.
% and $\chi^n(p||q) = (\frac{p}{q} - 1)^n$ is the Pearson-Vajda $\chi$-divergence of order $n$.
Recognize that $\int \datasetPolicy(a|s) (\frac{\pi(a|s)}{\datasetPolicy(a|s)} - 1)^n \mathrm{d}a = \chiAny{n}{\pi(\cdot|s)}{\datasetPolicy(\cdot|s)}$ is the Pearson-Vajda $\chi^n$ divergence.
\end{lemma}

For the second part of Theorem \ref{thm:reg_solution}, we need to show (i) the policy expression when $N=2$ and (ii) that when $N>5$ the solution does not have an analytic expression.
To show (i), we study regularization with the sole, general $\chi^n$.
Again let us write out the Lagrangian similar to Eq.(\ref{eq:f_lagrangian}):
\begin{align*}
\mathcal{L}(\pi, \alpha, \beta) = 
&\sum_s d^{\datasetPolicy}(s)\sum_{a}\AdaRectBracket{\pi(a|s)
Q(s,a)
  -
  \tau\, 
  % \AdaBracket{ \frac{\pi(a|s)}{\mu(a|s)}  - 1}^N
  \frac{\AdaBracket{\pi(a| s)-\mu(a| s)}^{N}}{\mu(a| s)^{\,N-1}}
}\\
&
-\sum_s d^{\datasetPolicy}(s)
\!\AdaRectBracket{\alpha(s)\AdaBracket{\sum_{a}\pi(a| s)-1}
-\sum_{a}\beta(a|s)\,\pi(a|s)}.
\end{align*}
where $d^{\datasetPolicy}, \alpha$ and $\beta$ are carry the same meaning as Eq.(\ref{eq:f_lagrangian}).
The KKT conditions are now:
\begin{align*}
&\text{Primal feasibility:}
\quad  \sum_{a}\pi(a| s)=1,
\quad \pi(a| s)\ge0, \\[3pt]
&\text{Dual feasibility:}
\quad  \beta(a|s)\ge 0 , \\[3pt]
&\text{Stationarity:}
\quad 
\frac{\partial\mathcal{L}}{\partial \pi(a| s)}
= Q(s,a)
-\tau\,\frac{N \AdaBracket{\pi(a| s)-\mu(a| s)}^{N-1}}
            {\mu(a| s)^{N-1}}-\alpha(s) + \beta(a|s)=0,\\[3pt]
&\text{Complementarity:}
\quad  \beta(a|s)\,\pi(a| s)=0 
\end{align*}
Following a similar procedure, we can obtain
\begin{align*}
Q(s,a) - \alpha(s) &= 
\tau\,\frac{N\,\AdaBracket{\pi(a| s)-\mu(a| s)}^{N-1}}
          {\mu(a| s)^{\,N-1}} \\
          &\Rightarrow
\AdaBracket{\pi(a|s) - \mu(a|s)}^{N-1}
=  \mu(a| s)^{N-1}  \frac{\AdaBracket{Q(s,a)-\alpha(s)}}{N\tau} .\\
&\Rightarrow \pi^*(a|s)  =  \mu(a|s) \AdaRectBracket{1 + \AdaBracket{ \frac{{Q(s,a)-\alpha(s)}}{N\tau}}^{\frac{1}{N-1}}}_{+} \,\,,
\end{align*}
where $\alpha(s)$ is the normalization constant ensuring $\sum_{a}\pi^*(a|s) = 1$. 
When $N=2$, this becomes 
\begin{align*}
    \pi^*(a|s)  &=  \mu(a|s) \AdaRectBracket{1 +  \frac{{Q(s,a)-\alpha(s)}}{2\tau}}_{+} 
\end{align*}
by redefining $\tau' = 2\tau$ we conclude the proof of (i).

Now let us consider the case where we have $\chi^2$ and $\chi^3$ appearing together, all other KKT conditions remain the same except for the stationarity:

\begin{align*}
\frac{\partial\mathcal{L}}{\partial \pi(a|s)}
=
Q(s,a) 
&-  \,2\tau \, \frac{f^{(2)}(1)}{2!} \frac{{\pi(a|s)-\mu(a|s)}}{\mu(a|s)}\\
&- \, 3\tau \, \frac{f^{(3)}(1)}{3!}\AdaBracket{\frac{\AdaBracket{\pi(a|s)-\mu(a|s)}}{\mu(a|s)} }^2 -\alpha(s) -\beta(a|s)=0,
% Q(s,a)-\alpha
% &= 2\tau \frac{f^{(2)}(1)}{2!} \frac{\pi(a|s)-\mu(a|s)}{\mu(a|s)} +
% 3\tau \frac{f^{(3)}(1)}{3!} \AdaBracket{\frac{{\pi(a|s)-\mu(a|s)}}{\mu(a|s)}}^{2}.
\end{align*}
Let us define
\begin{align*}
W(a|s) &:= \frac{\pi(a|s)-\mu(a|s)}{\mu(a|s)},  \quad \tau_2 := 2\tau \, \frac{f^{(2)}(1)}{2!} , \quad \tau_3 := 3\tau \, \frac{f^{(3)}(1)}{3!}\\[3pt]
&\Rightarrow\quad
W(a|s) = \frac{-\tau_2 + \sqrt{\tau_2^{2} + 4\tau_3 \AdaBracket{Q(s,a) - \alpha(s)}}}{2\tau_3}\\[3pt]
& \Rightarrow \pi^*(a|s) = \mu(a|s)\,
\AdaRectBracket{
  1+\frac{-\tau_2 + \sqrt{\tau_2^{2} + 4\tau_3\,\AdaBracket{Q(s,a)-\alpha}}}
       {2\tau_3} }_{+}.
\end{align*}
Though the reciprocal term becomes more complex, the role it plays still lies in determining the threshold for truncating actions.

Now similarly for $N=4$, we have $\chi^2$ to $\chi^4$ appearing together:
\begin{align*}
\frac{\partial\mathcal{L}}{\partial \pi(a|s)} = Q(s,a)
&- 2\tau \frac{f^{(2)}(1)}{2!}\frac{\pi(a|s)-\mu(a|s)}{\mu(a|s)} \\
&- 3\tau \frac{f^{(3)}(1)}{3!}\AdaBracket{\frac{\pi(a|s)-\mu(a|s)}{\mu(a|s)}}^2 \\
&- 4\tau \frac{f^{(4)}(1)}{4!}\AdaBracket{\frac{\pi(a|s)-\mu(a|s)}{\mu(a|s)}}^3
-\alpha(s)+\beta(a|s)=0.
\end{align*}
And again let us define:
\begin{align*}
W(a|s) &:= \frac{\pi(a|s)-\mu(a|s)}{\mu(a|s)},\\
\tau_2 &:= 2\tau \frac{f^{(2)}(1)}{2!},\qquad
\tau_3 := 3\tau \frac{f^{(3)}(1)}{3!},\qquad
\tau_4 := 4\tau \frac{f^{(4)}(1)}{4!}.
\end{align*}
Then for actions with $\pi^*(a|s)>0$, complementarity gives $\beta(a|s)=0$, and hence
\begin{align*}
&Q(s,a)-\alpha(s) = \tau_2 W(a|s) + \tau_3 W(a|s)^2 + \tau_4 W(a|s)^3.\\
\Rightarrow \quad&\tau_4 W(a|s)^3 + \tau_3 W(a|s)^2 + \tau_2 W(a|s) - \AdaBracket{Q(s,a)-\alpha(s)} = 0.
\end{align*}
After some algebra we obtain
\begin{align*}
&W(a|s) = \frac{1}{3\tau_4} \AdaRectBracket{-\tau_3 + \sqrt[3]{\frac{R+\sqrt{R^2+4B^3}}{2}} + \sqrt[3]{\frac{R-\sqrt{R^2+4B^3}}{2}}}, \\
\text{where }\,\, &B:=3\tau_4\tau_2-\tau_3^2,
\qquad
R:=27\tau_4^2\AdaBracket{Q(s,a)-\alpha(s)}+9\tau_4\tau_3\tau_2-2\tau_3^3.
\end{align*}
Therefore, the final solution is
\begin{align*}
\pi^*(a|s) = \mu(a|s) \AdaRectBracket{
1+ \frac{1}{3\tau_4} \AdaBracket{ -\tau_3 + \sqrt[3]{\frac{R+\sqrt{R^2+4B^3}}{2}} + \sqrt[3]{\frac{R \sqrt{R^2+4B^3}}{2}}}}_{+}.
\end{align*}
It is visible that though this solution still takes the form $\mu(a|s)\AdaRectBracket{1 + Z(s,a)}_{+}$, where $Z$ contains all other terms in the bracket.
The series $\sum_{n=2}^{N} {f^{(n)}(1)} \chi^n / n!$ is an $N$-th order polynomial in the policy ratio.
Therefore, for $N\geq 5$, by the famous Abel-Ruffini theorem \citep{Ramond-2022-AbelRuffni} we conclude that it is impossible to have any analytic solution.

\subsection{Proof of Theorem \ref{thm:cond_symm}}\label{apdx:conditional_symmetry}

The numerical instability of vanilla symmetric divergences comes from that the policy ratio is flipped.
Therefore, when either of the two policies becomes too small, there will be one side explodes. 
We address this again by the Taylor expansion.
However, a full Taylor expansion involves purely powers of policy ratio which can be numerically unstable for minimization.
Instead, we draw a key observation from Definition \ref{def:wider_symm} that the symmetric divergences can be decoupled into two interdependent terms $t\ln t$ and $g(t)$.
\begin{align}
        &\mathcal{L}(\theta) := \expectation{s\sim\mathcal{D}}{\lossD\AdaBracket{\regPi(\cdot|s) || {\pi_{\theta}(\cdot|s) } }} \nonumber \\
        &= \underbrace{\expectation{s\sim\mathcal{D}}{\KLany{\regPi(\cdot|s)}{\pi_{\theta}(\cdot|s)}}}_{t\ln t} \, + \, \underbrace{\expectation{s\sim\mathcal{D}\!}{\int\! \pi_{\theta}(a|s) \,\, g\!\AdaBracket{\frac{\regPi(a|s)}{\pi_{\theta}(a|s)}} \mathrm{d}a}}_\text{conditional symmetry} .
                % &= \expectation{s\sim\mathcal{D}}{\KLany{\regPi(\cdot|s)}{\pi_{\theta}(\cdot|s)}}\, + \, \expectation{s\sim\mathcal{D}\!}{\int\! \pi_{\theta}(a|s) \,\, g\!\AdaBracket{\frac{\regPi(a|s)}{\pi_{\theta}(a|s)}} \mathrm{d}a},
        \label{eq:lossD_f}
\end{align}
We can decompose the first term as follows:
\begin{align*}
    \KLany{\regPi(\cdot|s)}{\pi_{\theta}(\cdot|s)} &= \expectation{a\sim\regPi}{\ln\regPi(a|s) - \ln\pi_\theta(a|s)}\\
    &=  \expectation{a\sim\regPi}{ - \ln\pi_\theta(a|s)}\\
    &=  \expectation{a\sim\datasetPolicy}{- \AdaRectBracket{1 + \frac{Q(s,a) - V(s)}{\tau}}_{+}\ln\pi_\theta(a|s)}
\end{align*}
where the second step is because $\ln\regPi$ does not depend on the optimization variable $\theta$.
The last step is the result of substituting in the policy expression Eq. (\ref{eq:chi2_policy}): $\regPi(a|s)\propto\datasetPolicy(a|s)\exp\AdaBracket{\tau^{-1}\AdaBracket{Q(s,a)-V(s)}}$, leaving only $\datasetPolicy$ in the expectation subscript.
% Though this policy is intractable to directly compute due to its normalization constant, 
% it is known that by opting for the same KL as $\lossD$ in Eq. (\ref{eq:brpo_optimize}) yields
% \begin{align}
%     \pi_{\theta} = \argmin_{\theta}  \,\, \expectation{(s,a)\sim\mathcal{D}}{-\exp\AdaBracket{\frac{Q(s,a) - V(s)}{\tau}} \ln \pi_{\theta}(a|s)} .
%     \label{eq:forward_kl_loss}
% \end{align}
% The first term $t\ln t$ corresponds to the advantage regression in Eq.(\ref{eq:forward_kl_loss}).

We taylor-expand only the second term leveraging Lemma \ref{lemma:taylor}, obtaining 
\begin{align*}
    \int\! \pi_{\theta}(a|s) \,\, g\!\AdaBracket{\frac{\regPi(a|s)}{\pi_{\theta}(a|s)}} \mathrm{d}a = \int\! \pi_{\theta}(a|s) \,\, \sum_{n=2}^{N_\text{loss}} \frac{f^{(n)}(1)}{n!} \! \AdaBracket{\frac{\regPi(a|s)}{\pi_{\theta}(a|s)} \!-\!1 }^n\mathrm{d}a
\end{align*}
% For the $f$-divergence to be valid, the second term must also be convex. 
Note that $\regPi$ can be zero for some actions. 
For the vanilla symmetric divergences,  to ensure the second term is valid,  it is required that for actions sampled from $\pi_{\theta}$ the function $g$ cannot involve terms that flip the ratio e.g. $ -\ln t$ that destroys the validity.
The new objective does not have this restriction.

\begin{table*}[t!]
    \centering
    \resizebox{.99\textwidth}{!}{
    \begin{tabular}{p{3cm}p{4cm}p{3cm}p{4cm}p{3cm}}
    \toprule

% \multirow{6}{*}{Asymmetric}
  $\lossD$ & $f(t)$ & $g(t)$ & $g^{(n)}(1), (n\geq 2)$ & Series coefficient    \\ 
    % \cmidrule[0.25pt]{2-5}
    \midrule[0.25pt]
    %   & Forward KL  & $t\ln t$ & $(-1)^q (q-2)!$  & $\sum_{q=2}^{\infty} \frac{(-1)^q}{q(q-1)}$ \\ [3pt]
    % \cmidrule{2-5}
    % % Forward & $-\ln t$ & $(-1)^q (q-1)!$ &  $\sum_{q=2}^{\infty} \frac{(-1)^{q}(q-1)}{q}$ \\ 
    % % |rule
    % % Backward $\KLindex{k}{k+1}$ & $t\ln t$ & $(-1)^q (q-2)!$  & $\sum_{q=2}^{\infty} \frac{(-1)^{q}}{q}$ \\ 
    % % |rule
    % &Backward KL & $-\ln t$ & $(-1)^q (q-1)!$ &  $\sum_{q=2}^{\infty} \frac{(-1)^q}{q}$ \\ [3pt]
    % |rule[1pt]
    % \multirow{5}{*}{Symmetric}
    Jeffrey & $(t-1)\ln t$ & $-\ln t$ & $(-1)^n {(n-1)!}$  & $\sum_{n=2}^{\infty} \, (-1)^n \frac{1}{n} $ \\ [3pt]
% $\sum_{q=2}^{\infty} \! \frac{(-1)^q (q-1)}{q} \! \AdaBracket{1 + \frac{1}{q-1}}$
    % \cmidrule{2-5}
    \midrule
    Jensen-Shannon   &  $t\ln t - (1+t)\ln\frac{1+t}{2}$ &  
   $-(1+t)\ln\frac{1+t}{2}$  &$(-1)^n (n-2)! {\frac{1}{2^{n-1}}}$ & $\sum_{n=2}^{\infty}\frac{(-1)^n }{n(n-1)2^{n-1}}$ \\ [3pt]
    % \cmidrule{2-5}
    \midrule
    GAN Divergence   &  $t\ln t - (1+t)\ln\AdaBracket{1+t}$ &  $-(1+t)\ln (1+t)$
  &$(-1)^n (n-2)! \frac{1}{2^{n-1}}$ &  $\sum_{n=2}^{\infty}\frac{(-1)^n }{n(n-1)2^{n-1}}$ \\ [3pt]
    \bottomrule
    \end{tabular}}
    \caption{
      Symmetric divergences, their $f$ generators, conditional symmetry $g$, derivatives and Taylor expansion series coefficients. 
      JS and GAN share the same derivatives and series coefficients.
      }
      \label{table:approx_table}
  \end{table*}

Adding both terms, we obtain our final loss objective:
\begin{align*}
        \mathcal{L}(\theta) \!=\! 
        % &= \expectation{\substack{s\sim\mathcal{D}\\a\sim\pi_{\theta}}\!}{{ f_\text{ConSym}\!\AdaBracket{\frac{\regPi(a|s)}{\pi_{\theta}(a|s)}}  }} \\
        % &= \expectation{\substack{s\sim\mathcal{D}\\a\sim\pi_{\theta}}}{{ 
        % \sum_{n=2}^{\infty} \frac{f^{(n)}(1)}{n!}\AdaBracket{\frac{\regPi(a|s)}{\pi_{\theta}(a|s)}-1}^n  }} \\
        \expectation{(s,a)\sim\mathcal{D}\!}{-\AdaRectBracket{1 +\frac{Q(s,a) \!- \! V(s)}{\tau}}_{+} \ln \pi_{\theta}(a|s)} \!+ {\expectation{\substack{s\sim\mathcal{D}\\a\sim\pi_{\theta}\!}}{{ 
        \sum_{n=2}^{N_\text{loss}} \frac{f^{(n)}(1)}{n!} \! \AdaBracket{\frac{\regPi(a|s)}{\pi_{\theta}(a|s)} \!-\!1 }^n}}}.
\end{align*}
% leverage the Taylor expansion to expand the term $f_\text{ConSym}(t) = f(1) + \sum_{n=0}^{\infty}\AdaBracket{f^{(n)}(1)/n!}(t-1)^i$:
% \begin{align}
%     \begin{split}
%         \mathcal{L}_{\text{ConSym}}(\theta) 
%         % &= \expectation{\substack{s\sim\mathcal{D}\\a\sim\pi_{\theta}}\!}{{ f_\text{ConSym}\!\AdaBracket{\frac{\regPi(a|s)}{\pi_{\theta}(a|s)}}  }} \\
%         % &= \expectation{\substack{s\sim\mathcal{D}\\a\sim\pi_{\theta}}}{{ 
%         % \sum_{n=2}^{\infty} \frac{f^{(n)}(1)}{n!}\AdaBracket{\frac{\regPi(a|s)}{\pi_{\theta}(a|s)}-1}^n  }} \\
%         &= 
%         \expectation{\substack{s\sim\mathcal{D}\\a\sim\pi_{\theta}}}{{ 
%         \sum_{n=2}^{\infty} \frac{f^{(n)}(1)}{n!}\chiAny{n}{\policydot{\regPi}}{\policydot{\pi_{\theta}} }  }},
%     \end{split}
%     \label{eq:taylor_loss}
% \end{align}
% where we let $z=1$. 
% $f^{(n)}$ is the $n$-th derivative and $f^{(0)}=f$.
% Note that the series starts from $n=2$ since $f(1)=0$ and $\chi^n(\cdot||\cdot)=0$ when $n=1$.
% The last line is because of the classic result of approximating any bounded and smooth $f$-divergence with higher order $\chi^n$ divergences \citep{Barnett2002-approximateFdivTaylorSeries,Nielsen2013-chiApproxFdiv}.
Table \ref{table:approx_table} summarizes the series coefficients.
Compared to full expansion, expanding only the conditional symmetry part has an additional benefit: for large $n$ with high order policy ratio,
% $\AdaBracket{\frac{\regPi}{\pi_{\theta}} - 1}^n$,
its coefficient decays quickly towards zero and lowers the importance of higher order terms.

\subsection{Proof of Theorem \ref{thm:error_bound}}\label{apdx:error_bound}

To prove Theorem \ref{thm:error_bound}, we require that the policy ratio is clipped to $[1-\epsilon, 1+\epsilon], \,\, \epsilon > 0$.
This is because Lemma \ref{lemma:taylor} assumed the Taylor expansion was around $t=1$.
Therefore, the policy ratio should stay in the neighborhood of $1$ for the series to converge.
% This justifies clipping to $[1-\epsilon, 1+\epsilon], \,\, \epsilon > 0$.
As such, Taylor expansion provides an interesting interpretation to the proximal policy optimization (PPO) style clipping \citep{schulman2017proximal,Vaswani2022-generalSurrogate-FMAPG,zhuang2023-behaviorPPO}: 
it is the convergence radius for regularized BRPO.
% From a practical viewpoint, clipping the policy ratio helps stabilizing and improving learning.
% For the latter, we truncate it to the first $k$ terms.
% Let us denote the $k$-term truncation of Eq.(\ref{eq:taylor_loss}) as $\widehat{\mathcal{L}}_\text{ConSym}^k(\theta)$.
% We can show that minimizing the $k$-term approximation is not far from minimizing its exact counterpart, in that they have bounded error:
% Moreover, we can show that the distance from the clipped truncated series to its exact counterpart is upper-bounded:

We follow \citep[Theorem 1]{Barnett2002-approximateFdivTaylorSeries} in proving this result.
We start with the following Taylor representation with the integral remainder:
\begin{align*}
    f(t) = f(z) + \sum_{n=0}^{N} \frac{\AdaBracket{t-z}^n}{n!} f^{(n)}(z) + \frac{1}{N!}\int^{t}_{z} (t-z)^N f^{(N+1)}(a) \, \mathrm{d}a.
\end{align*}
Specifically by (2.4) of \citep{Barnett2002-approximateFdivTaylorSeries} we have
\begin{align*}
    \adaAbsolute{f(t) - f(z) - \sum_{n=0}^{N} \frac{\AdaBracket{t-z}^n}{n!} f^{(n)}(z)} &\leq \frac{1}{N!} \adaAbsolute{\int^{t}_{z} |t-z|^N \adaAbsolute{f^{(N+1)}(a)} \, \mathrm{d}a} \\
    & \leq \frac{1}{N!} \sup_{a\in [1-\epsilon, 1+\epsilon]}\adaAbsolute{f^{(N+1)}(a)}\adaAbsolute{\int^{t}_{z} |t-a|^N \mathrm{d}a} \\
    & =\frac{1}{(N+1)! } \mynorm{f^{(N+1)}}{\infty} |t-z|^{N+1} \\
    & = \frac{1}{(N+1)! } \mynorm{f^{(N+1)}}{\infty} \epsilon^{N+1},
\end{align*}
where in the last equation we let $t = \regPi/\pi_{\theta}$ clipped to $1+\epsilon$ and $z=1$.
Now we can repeat the same procedure for $t = 1-\epsilon$.
Since states are sampled from the dataset randomly, 
% and take an expectation over $s\sim\mathcal{D}$:
we have
\begin{align*}
    \expectation{s\sim\mathcal{D}}{\frac{2\epsilon^{N+1}}{(N+1)! } \mynorm{f^{(N+1)}}{\infty} } = \sum_{s} \frac{1}{|\mathcal{D}|}\frac{2\epsilon^{N+1}}{(N+1)! } \mynorm{f^{(N+1)}}{\infty} = \frac{2\epsilon^{N+1}}{(N+1)! } \mynorm{f^{(N+1)}}{\infty}.
\end{align*}
We conclude the proof of Theorem \ref{thm:error_bound} by changing $f$ to $g$.

% \subsection{Symmetric Divergences Do not Permit Closed-form Policies}\label{apdx:no_closed_form}

\subsection{Existing Characterizations of $f$-regularized Policy}\label{apdx:review_no_closed}

We review related work on characterizing the solution of \fdiv regularization. 
They are mainly two ways for characterization, which we discuss in detail below.

\begin{minipage}{0.48\textwidth}
    \begin{tcolorbox}[
  title=Regularization Characterization,
  boxrule=0.5pt, % Frame thickness
  arc=4pt, % Rounded corners
  % fonttitle=\bfseries, % Bold title font
  enhanced, % Use enhanced drawing capabilities
  colback=gray!20, % Light gray background
  colframe=gray!50, % Medium gray frame
  coltitle=black, % Black title text
  ]
\textbf{(R1)} $\regPi(a|s) > 0 \Rightarrow \datasetPolicy(a|s)>0$; \\
\textbf{(R2)} $ h_f(t) := t f(t)$ is strictly convex; \\
\textbf{(R3)} $f(t)$ is continuously differentiable.\\
\textbf{Result: }
\begin{center}
    % $\regPi(a|s) \propto \max\AdaCurlyBracket{0, (h_f')^{\raisebox{0.5ex}{$\scriptstyle \!-1$}} \!\AdaBracket{\frac{Q(s,a)}{\tau}}}$.
        $\regPi(a|s) \propto \AdaRectBracket{ \,(h_f')^{\raisebox{0.5ex}{$\scriptstyle \!-1$}} \!\AdaBracket{\frac{Q(s,a)}{\tau}} \,}_{+}$.
\end{center}
\end{tcolorbox}
\end{minipage}
\hfill
\begin{minipage}{0.48\textwidth}
\begin{tcolorbox}[
  title=DICE Characterization,
  boxrule=0.5pt, % Frame thickness
  arc=4pt, % Rounded corners
  enhanced, % Use enhanced drawing capabilities
  colback=gray!20, % Light gray background
  colframe=gray!50, % Medium gray frame
  coltitle=black, % Black title text
  ]
\textbf{(D1)} $d^{\regPi}(s,a) > 0 \Rightarrow d^{\datasetPolicy}(s,a) > 0$; \\
\textbf{(D2)} $f(t)$ is strictly convex; \\
\textbf{(D3)} $f(t)$ is continuously  differentiable.\\
\textbf{Result: }
\begin{center}
     % $\frac{d^{\regPi}(s,a)}{d^{\datasetPolicy}(s,a)} \propto \max\AdaCurlyBracket{0, (f')^{\raisebox{0.5ex}{$\scriptstyle \!-1$}} \!\AdaBracket{\frac{Q(s,a)}{\tau}}}$.
          $\frac{d^{\regPi}(s,a)}{d^{\datasetPolicy}(s,a)} \propto \AdaRectBracket{ \, (f')^{\raisebox{0.5ex}{$\scriptstyle \!-1$}} \!\AdaBracket{\frac{Q(s,a)}{\tau}} \,}_{+}$.
\end{center}
\end{tcolorbox}
\end{minipage}

\subsubsection{Regularization Characterization. }

We call the first class Regularization Characterization as they exactly studied Eq. (\ref{eq:brpo})
\citep{Li2019-regularizedSparse,Xu2023-OfflineNoODDAlphaDiv}.
% \todo[inline,bordercolor=gray, backgroundcolor=gray!10]{\textbf{(D1)} $f(1) = 0$; 
% \textbf{(D2)} the function $h_f(t) := t f(t)$ is strictly convex; 
% \textbf{(D3)} $f(t)$ is differentiable.}
% \textbf{(D1)} $f(1) = 0$; \\
% \textbf{(D2)} the function $h_f(t) := t f(t)$ is strictly convex; \\
% \textbf{(D3)} $f(t)$ is differentiable.\\
Here, $\propto$ indicates \emph{proportional to} up to a constant not depending on actions.
Assumptions (R2) does not hold for symmetric divergences in general.

\textbf{Jeffrey's divergence. } $D_\text{Jeffrey}(\regPi ||\pi_{\theta} ) = \KLany{\regPi}{\pi_{\theta}} + \KLany{\pi_{\theta}}{\regPi}$ is induced by $f(t) = (t-1)\ln t$.
We see that $h_f(t) = (t^2-t) \ln t$, and therefore $h'_f(t) = (2t-1)\ln t + t - 1$;  $h''_f(t) = 2\ln t + 3 - \frac{1}{t}$, which can be negative and in turn indicates that $h_f$ is not strictly convex.
Therefore, Jeffrey's divergence does not satisfy their Assumption (R2).
% This is also visible from the fact that $h_f'(t)$ involves  $t\ln t$.
% As a corollary, the asymmetric reverse KL induced by $f(t) := -\ln t$ fails to pass as well since its  $h_f''(t) = 2\ln t + 3$ can also be negative.

% \begin{tcolorbox}[
%   title=DICE Characterization,
%   boxrule=0.5pt, % Frame thickness
%   arc=4pt, % Rounded corners
%   breakable, % Allow the box to break across pages if needed
%   left=5pt, % Left padding
%   right=5pt, % Right padding
%   top=5pt, % Top padding
%   bottom=5pt, % Bottom padding
%   before skip=5pt, % Space before the box
%   after skip=5pt, % Space after the box]
%   ]
% \textbf{(D1)} $f(1) = 0$; \\
% \textbf{(D2)} the function $f(t)$ is strictly convex; \\
% \textbf{(D3)} $f(t)$ is differentiable.
% \begin{center}
%     \textbf{Result: } $\frac{\regPi(a|s)}{\datasetPolicy(a|s)} \propto \max\AdaCurlyBracket{0, (f')^{\raisebox{0.5ex}{$\scriptstyle \!-1$}} \!\AdaBracket{\frac{Q(s,a)}{\tau}}}$.
% \end{center}
% \end{tcolorbox}
% This indicates a limited applicable range of their method.

\textbf{Jensen-Shannon Divergence. }
Recall the Jensen-Shannon divergence is defined by 
\begin{align*}
    &f(t) := t\ln t - (t+1)\ln\frac{t+1}{2}.
\end{align*}
We examine Assumption (R2) of regularization characterization \citep{Li2019-regularizedSparse,Xu2023-OfflineNoODDAlphaDiv}:
\begin{align*}
    &h_f(t) := t f(t) =  t^2\ln t - t^2\ln\frac{t+1}{2}-t\ln\frac{t+1}{2},\\
    &\Rightarrow h'_f(t) = 2t\ln t + t - 2t\ln\left(\frac{t+1}{2}\right) - \frac{t^2}{t+1} - \ln\left(\frac{t+1}{2}\right) - \frac{t}{t+1}, \\
    & \Rightarrow h''_f(t) = 2\ln\left(\frac{2t}{t+1}\right)+\frac{1}{t+1}.
\end{align*}
suggesting that $h_f(t)$ is not a convex function and does not meet their Assumption (R2).

\textbf{GAN Divergence. }
From Table \ref{tab:f-divergences} the GAN divergence is defined by 
\begin{align*}
    f(t)=t \ln t - (t+1)\ln(t+1).
\end{align*}
Again we focus on its second assumption:
\begin{align*}
&h_f(t)=t^2 \ln t - t^2\ln(t+1) - t \ln(t+1) \\    
\Rightarrow \,\, &h'_f(t) = 2t\ln\AdaBracket{\frac{t}{t+1}} - \ln(t+1)\\
\Rightarrow \,\, &h''_f(t) = 2\ln\AdaBracket{\frac{t}{t+1}} + \frac{1}{t+1}.
\end{align*}
$h''_f(t)$ can be negative, therefore Assumption (R2) is not satisfied.

\subsubsection{DICE Characterization}

% The DICE characterization directly works on the generator function, and derive the policy solution using $\max\AdaCurlyBracket{0, (f')^{-1}\!\AdaBracket{\frac{A(s,a)}{\tau}}}$, where $A$ is the advantage function \citep{lee2021-OptidiceOffline}.

% \textbf{DICE Characterization. }
DIstribution Correction Estimation (DICE) methods estimate stationary distribution ratios that correct the discrepancy between the data distribution and the optimal policy’s stationary distribution \citep{Nachum2019-DualDICE,Nachum2020-RLFenchelRockafellar}.

In the offline context, the optimal solution is the ratio between the stationary distributions ${d^{\regPi}(s,a)} / {d^{\datasetPolicy}(s,a)}$ \citep{lee2021-OptidiceOffline,Mao2024-ODICE-OrthogonalDice}.
The optimal policy can then be uniquely identified by $\regPi(a|s) = {d^{\regPi}(s,a)}/{\sum_{b} d^{\regPi}(s,b)}$ \citep{Puterman1994}.
Assumptions (D1)-(D3) can be satisfied by the symmetric divergences in Table \ref{tab:f-divergences}.
However, the issue lies in that $(f')^{-1}$ in general does not have a closed-form expression.

\textbf{Jeffrey's divergence. } $f(t) = (t-1)\ln t$, the inverse function of $f'(t) = \ln t + 1 - \frac{1}{t}$ involves the Lambert W function which does not have an analytic expression \citep{Nowozin2016-fGAN}.

\textbf{Jensen-Shannon Divergence. }
The generator function of Jensen-Shannon divergence is:
\begin{align*}
    &f(t) := t\ln t - (t+1)\ln\frac{t+1}{2}
    \quad \Rightarrow \quad f'(t) = \ln t - \ln \frac{t+1}{2},\\
    & \Rightarrow \regPi(a|s) \propto \max\AdaCurlyBracket{0, \,\,\AdaBracket{f'}^{-1} \!\AdaBracket{\frac{Q(s,a)}{\tau}} }= \max\AdaCurlyBracket{0,\,\, \frac{\exp\AdaBracket{\frac{Q(s,a)}{\tau}}}{2 - \exp\AdaBracket{\frac{Q(s,a)}{\tau}}}}.
\end{align*}
To make sure the policy is a valid distribution, we need to find the normalization constant. However, the integral of $(f')^{-1}$ diverges to infinity, suggesting that no such normalization constant exists.

\section{Additional Results}\label{apdx:additional}

\subsection{Didactic Examples}\label{apdx:toy_examples}

This section provides details of the examples shown in the main text.
The first example, involving outliers under critic misspecification, further motivates the importance of symmetric regularization, while the second provides details of the divergence-minimization example.

\subsubsection{Rare Outlier Under Critic Misspecification }
Consider a three-action bandit with behavior policy
\[
\datasetPolicy = (0.60,\,0.39,\,0.01),
\]
with true values $Q$ and misspecified critic $\hat{Q}$:
\[
Q_{\mathrm{true}} = (1.0,\,0.95,\,-6.0), \qquad \hat Q = (1.0,\,0.95,\,-6.0 + b).
\]
% and misspecified critic
% \[
% \hat Q = (1.0,\,0.95,\,-6.0 + b).
% \]
The third action is rare in the offline dataset and is poor.
Assume that the critic receives a positive bias $b$ on that action, which is common in offline RL \citep{Fujimoto18-addressingApproximationError}.
Sweeping $b \in [0,10]$ shows that the solution under $D_\mathrm{KL}(\datasetPolicy\|\pi)$ is the first to concentrate on the spurious rare action, while the Jeffreys-regularized solution is the most stable among the three compared methods. At $b=8.0$, we have
% This example targets a different pathology. The issue is no longer a refusal to move toward a rare good action. Instead, the issue is whether the policy can be \emph{too eager} to form a rare mode when the critic spuriously exaggerates its value.
% \begin{example}[Rare outlier]
% Consider \eqref{eq:brpo-original} with the instance above. In the sweep $b \in [0,10]$, the optimizer under $\mathrm{KL}(\mu\|\pi)$ is the first to concentrate on the spurious rare action, while the Jeffreys-regularized optimizer is the most stable among the three compared methods. At the representative bias $b=8.0$,
\[
\pi_{\mathrm{KL}(\pi\|\mu)}(a_{\mathrm{outlier}})=0.072,\qquad
\pi_{\mathrm{KL}(\mu\|\pi)}(a_{\mathrm{outlier}})=0.519,\qquad
\pi_\mathrm{JS}(a_{\mathrm{outlier}})=0.036,
\]
and the corresponding true values are
\[
\mathbb{E}_{a \sim \pi_{\mathrm{KL}(\pi||\datasetPolicy)}}[Q_{\mathrm{true}}(a)] = 0.479,
\]
\[
\mathbb{E}_{a \sim \pi_{\mathrm{KL}(\datasetPolicy||\pi)}}[Q_{\mathrm{true}}(a)] = -2.641,\qquad
\mathbb{E}_{a \sim \pi_\mathrm{JS}}[Q_{\mathrm{true}}(a)] = 0.733.
\]
% \end{example}
\begin{figure}[t]
    \centering
    \includegraphics[width=0.95\textwidth]{figs/toy_examples/outlier_trap.pdf}
    \caption{
    Symmetric regularization helps in the case where rare actions have bias. 
    (Left) probability mass assigned to the spurious rare action. 
    (Right) Resulting true values obtained by different divergence regularizers.}
    \label{fig:outlier-trap}
\end{figure}

In this example, reverse KL is fragile: it concentrates on an action that is rare under the behavior policy whenever that action is assigned a favorable value, which can lead to spurious critic values; see Figure \ref{fig:outlier-trap}. 
% This leads to critic hallucination. 
By contrast, Jeffreys helps because it reintroduces the forward-KL pressure against assigning additional mass to a rare action. 
% That extra symmetry is helpful in this instance.

% There is also an important caveat. This example should not be read as ``any symmetric divergence solves outlier chasing.'' In preliminary sweeps for this configuration, Jensen--Shannon was too permissive and could also collapse to the outlier. 
% When reverse-KL-like behavior is too eager to create a rare spurious mode, a symmetric divergence with a sufficiently strong forward component, such as Jeffreys, can be materially more robust.
% \end{interpretation}

\subsubsection{Boundary Fitting Under Action Clipping }
% \paragraph{Setting.}
We set the action range as $[-1,1]$, which is common to many applications. We use an unconstrained Gaussian 
$\pi_\theta = \mathcal{N}(m,s^2)$.
The target policy is a truncated mixture concentrated near the left boundary:
\[
\pi^\star(a) \propto 0.8 \, \mathcal{TN}(-0.96,\,0.03^2) + 0.2 \, \mathcal{TN}(-0.72,\,0.09^2),
\qquad a \in [-1,1].
\]
$\pi_\theta$ is obtained by minimizing KL or JS.
Actions outside the range will be clipped.
We design the reward function to be
\[
r(a)=\exp\!\left(-\frac{(a+0.96)^2}{2(0.025)^2}\right)
+ 0.6 \exp\!\left(-\frac{(a+0.72)^2}{2(0.08)^2}\right).
\]

% \begin{example}[Boundary leakage]
The forward KL divergence produces a much wider Gaussian distribution than the Jensen-Shannon solution. As a result, a significantly larger fraction of the probability mass falls outside the allowed action range. 
The fitted parameters and induced clipped rewards are
\[
(m_{\mathrm{KL}},s_{\mathrm{KL}}) = (-0.907,0.105),\qquad
(m_{\mathrm{JS}},s_{\mathrm{JS}}) = (-0.953,0.025).
\]
The corresponding off-support masses and clipped rewards are
\[
\text{off-support}_{\mathrm{KL}}=18.8\%,\qquad
\text{off-support}_{\mathrm{JS}}=3.3\%,
\]
\[
\mathbb{E}[r(\mathrm{clip}(a,-1,1))]_{\mathrm{KL}}=0.383,\qquad
\mathbb{E}[r(\mathrm{clip}(a,-1,1))]_{\mathrm{JS}}=0.703.
\]
% \end{example}

% \begin{figure}[t]
%     \centering
%     \includegraphics[width=\textwidth]{figs/toy_examples/boundary_example.pdf}
%     \caption{Boundary-fitting instance. The dashed line in the left panel marks the clipping boundary at $a=-1$. The two right panels separately report the fraction of Gaussian mass outside the legal range and the expected reward after clipping.}
%     \label{fig:boundary}
% \end{figure}

\begin{table}[t]
    \centering
    \caption{Boundary fitting summary statistics.}
    \label{tab:boundary}
    \begin{tabular}{lcccc}
        \toprule
        Method & Mean & Std. dev. & Off-support mass (\%) & Clipped reward \\
        \midrule
        $\mathrm{KL}(\pi^\star\|\pi_\theta)$ & -0.907 & 0.105 & 18.8 & 0.383 \\
        $\mathrm{JS}(\pi^\star ||\pi_\theta)$ & -0.953 & 0.025 & 3.3 & 0.703 \\
        \bottomrule
    \end{tabular}
\end{table}

The result can be seen in Table \ref{tab:boundary}.
Forward KL is mass-covering and attempts to cover the entire target support.
On the other hand, it does not have any mechanism to prevent extra probability mass falling into impossible actions. 
In this case, actions outside the range are clipped and therefore alter the resulting policy distribution. 
Jensen-Shannon penalizes the mismatch from both sides, suppressing probability-mass leakage and thereby improving the post-clipping reward.
This coincides with recent findings in \citep{Zhu2025-qExpPolicy,Lee2025-truncatedGaussianPolicy}.
% \end{interpretation}

\subsubsection{Mismatched Geometry Induced by Different $\regD$ and $\lossD$}\label{apdx:mismatched_geometry}

Consider a three-action bandit with the following target:
\[
\pi^*_{\mathrm{Reg}} = (0.05,\,0.70,\,0.25).
\]
We restrict the actor class to
\[
\pi_A = (0.75,\,0.15,\,0.1),
\qquad
\pi_B = (0.2,\,0.05,\,0.75).
\]
If the regularizer is the forward KL $D_\mathrm{KL}(\pi\|\pi^*_{\mathrm{Reg}})$, then the matched restricted update also minimizes forward KL, as can be seen by:
\[
D_\mathrm{KL}(\pi_A\|\pi^*_{\mathrm{Reg}})=1.708
\;>\;
D_\mathrm{KL}(\pi_B\|\pi^*_{\mathrm{Reg}})=0.969,
\]
so the exact restricted update chooses $\pi_B$. 
If instead the reverse KL $D_\mathrm{KL}(\pi^*_{\mathrm{Reg}}||\pi)$ is minimized, the inequality flips:
\[
D_\mathrm{KL}(\pi^*_{\mathrm{Reg}}\|\pi_A)=1.172
\;<\;
D_\mathrm{KL}(\pi^*_{\mathrm{Reg}}\|\pi_B)=1.503,
\]
and $\pi_A$ is chosen instead. 
In this case, the resulting gap is
\[
\varepsilon_{\mathrm{gap}}=1.708-0.969=0.739.
\]
Therefore, it is often suboptimal to use a different $\lossD$ than $\regD$, even in the case of forward versus reverse KL.
The Jensen-Shannon case typically exhibits a larger gap than the reverse-KL case.
% This is the sharpest warning against matching only the divergence family name. For Bregman regularizers, orientation is part of the geometry. Using the wrong orientation can flip the selected actor even when the loss is still called KL.

% Figure~\ref{fig:projection-mismatch} reports the numerical values for the three mismatch cases. In the left panel, a KL-regularized target fitted with Jensen--Shannon on a two-candidate class incurs an excess gap of $0.080$. In the middle panel, merely reversing the KL orientation incurs a much larger gap of $0.739$. In the right panel, the actor class is a width-1 bottleneck neural manifold; the matched projection selects $u_{\mathrm{match}}=0.293$, while Jensen--Shannon fitting selects $u_{\mathrm{fit}}=-0.021$, producing an excess gap of $0.035$. The neural example is especially important for practice: once the actor is a narrow network, fitting is unavoidably a restricted-function-class projection problem.

\subsection{Distribution Matching}\label{sec:approx_mixture}

Every loss divergence should ideally lead to the same optimal solution.
As a sanity check, we first verify that our Taylor-expanded loss in Eq. (\ref{eq:taylor_loss}) is valid, in the sense that minimizing it produces results similar to those of existing divergence losses.
We follow the setting in \citep{Nowozin2016-fGAN} to learn a Gaussian $\pi_{\theta}$ with parameters $\theta = (\mu, \sigma)$ that approximates a univariate Gaussian mixture.
The target mixture is shown in Figure \ref{fig:approximate_mixture}.

\textbf{Setup.}
Learning is performed by minimizing the expanded objective in Eq. (\ref{eq:taylor_loss}) with $N_\text{loss}=5$.
The target policy is the mixture distribution, and $\theta$ parameterizes a two-layer neural network with 64 hidden units.
We minimize the objective by SGD with learning rate $0.001$, batch size $128$ for 1000 update steps.
The optimal Gaussian parameters $\mu, \sigma$ are obtained by numerical integration.
We compare them with the parameters obtained by minimizing our objective and the vanilla divergences.
% The learned $(\hat{\mu}_{g}, \hat{\sigma}_g)$ are compared to the best fit $(\mu^*, \sigma^*)$ obtained by numerical integration. 
%
% We also compare to $(\hat{\mu}, \hat{\sigma})$ learned by minimizing the original $f$-divergences.

\begin{minipage}{0.55\textwidth}
% \begin{table}[h!]
%     \centering
\resizebox{\textwidth}{!}{%
    \begin{tabular}{l l b a b a}
        \toprule
        & Method   & Jeffreys (values) & Jeffreys (abs. error) $\downarrow$ & JS (values) & JS (abs. error) $\downarrow$\\
        \midrule[0.25pt]
        \multirow{3}{*}{$D_f$} & Optimal        & $0.0159$ & $-$ & $0.0368$ & $-$ \\
        & Vanilla & $0.0157 $ & $0.0002$ & $0.0682$ & $0.0314$ \\
        & Ours & $0.0161$  & $0.0002$ & $0.0387$  & $0.0019$ \\
        \midrule[0.25pt]
        \multirow{3}{*}{$\mu$} & Optimal  & $0$ & $-$ & $0$ & $-$  \\
        & Vanilla & $0.0067$ &  $0.0067$ & $-0.0778$ & $0.0778$ \\
        & Ours & $-0.0166$ & $0.0166$ & $-0.0475$ & $0.0475$   \\
        \midrule[0.25pt]
        \multirow{3}{*}{$\sigma$ } & Optimal &  $2.2218$ & $-$ & $2.2868$ & $-$  \\
        & Vanilla &  $2.2396$ & $0.018$ & $4.5559$ & $2.2691$ \\
        & Ours &  $2.4926$ & $0.2707$ & $2.5438$ & $0.2570$  \\
        \bottomrule
    \end{tabular}
    % \begin{tabular}{l l c c c}
    %     \toprule
    %     & Methods & Forward KL  & Jeffrey & Jensen-Shannon  \\
    %     \midrule[0.25pt]
    %     \multirow{3}{*}{$D_f$} & Numerical Integration          & $0.0146$ & $0.0159$ & $0.0368$ \\
    %     & Original & $0.0143$ & $0.0155$ & $0.0682$ \\
    %     & Ours & $-$ & $0.0157$ & $0.0387$  \\
    %     \midrule[0.25pt]
    %     \multirow{3}{*}{$\mu$} & Numerical Integration & $1.762 \!\times\! 10^{-5}$ & $-3.304\!\times\! 10^{-5}$ & $-4.334\!\times\! 10^{-5}$  \\
    %     & Original & $-0.0030$ & $0.0067$ & $-0.0778$ \\
    %     & Ours & $-$ & $-0.0166$ & $-0.0475$    \\
    %     \midrule[0.25pt]
    %     \multirow{3}{*}{$\sigma$ } & Numerical Integration & $2.2358$ & $2.2218$ & $2.2868$   \\
    %     & Original & $2.2519$ & $2.2396$ & $4.5559$  \\
    %     & Ours & $-$ & $2.4926$ & $2.5438$  \\
    %     \bottomrule
    % \end{tabular}
    }
    \captionof{table}{Loss objectives $D_f$ and corresponding fit parameters $\mu, \sigma$. 
    Optimal: values given by numerical integration.
    % gives the optimal divergence values $D_f$ and parameters $\mu, \sigma$. 
    Vanilla: minimized the vanilla $f$-divergence per definition. 
    Ours: minimized our Taylor expansion loss for $N_\text{loss}=5$.
    Our JS loss yields a better fit than minimizing the vanilla $f$-divergence.
    }
    \label{tab:approx_result}
% \vspace{5pt}
\end{minipage}\hfill
% \begin{minipage}[t]{0.6\textwidth}
%     \centering
\begin{minipage}{0.425\textwidth}
        \includegraphics[width=\textwidth]{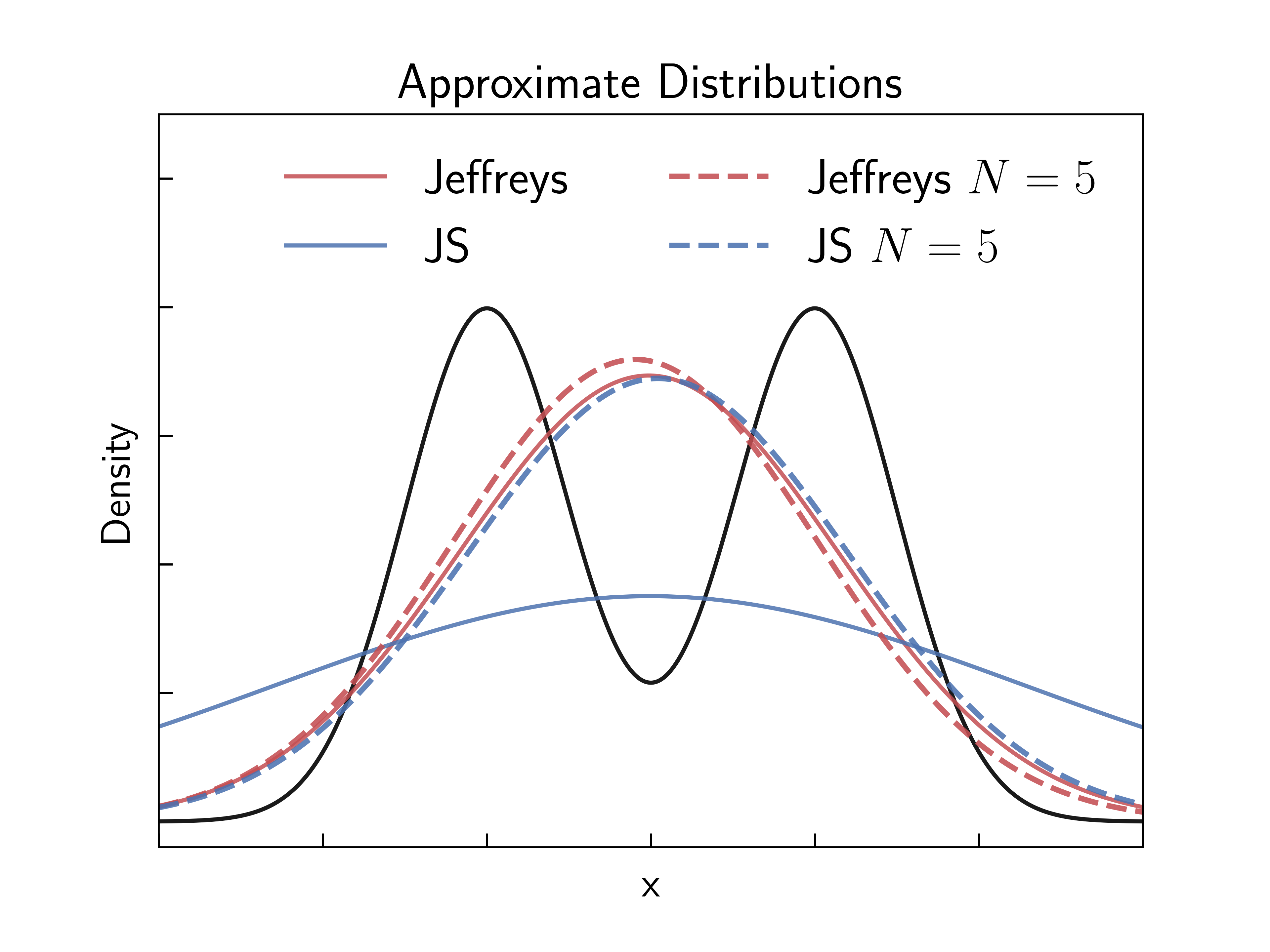}
        \captionof{figure}{
        Approximating a mixture of Gaussians (black) by minimizing vanilla divergence (solid) and S$f$-AC loss for $N_\text{loss}\!=\!5$ (dashed).
        Vanilla JS loss causes the Gaussian to lose track of optimal $\sigma^*$ given by numerical integration.
        }
        \vspace{-20pt}
        \label{fig:approximate_mixture}
        \vspace{1cm}
\end{minipage}
% \begin{minipage}{0.48\textwidth}
%         \includegraphics[width=\textwidth]{figs/conditional_symmetry_expansion.png}
%         % \captionof{figure}{Caption}
%         % \label{fig:enter-label}
% \end{minipage}
% \captionof{figure}{A single caption spanning two side-by-side figures illustrating some related concepts.}
%     \label{fig:widecaption}
% \end{minipage}

% as follows.
% \begin{eqnarray}
% & x &
%   \to \textrm{Linear(1,64)}
%   \to \textrm{Tanh}
%   \to \textrm{Linear(64,64)}
%   \to \textrm{Tanh}
%   \to \textrm{Linear(64,1)}
% \end{eqnarray}
%

\textbf{Results.}
Table~\ref{tab:approx_result} shows the divergence values and fitted Gaussian parameters obtained by numerical integration (Optimal), the vanilla objective (Vanilla), and our method (Ours).
% $D_f^*$ denotes that of the best fit, and $\widehat{D}_{f}$ that of original divergence.
The S$f$-AC loss is a reasonable objective and induces learned distributions (dashed curves) that are consistent with the numerical solution. 
Moreover, as can be seen in Figure \ref{fig:approximate_mixture}, minimizing the vanilla Jensen-Shannon (JS) divergence loss induces a much wider distribution (solid blue) with $\sigma = 4.556$.
The poor approximation coincides with the observation that exact symmetric divergence losses can lead to unstable policy behaviors \citep{Wang2024-beyondReverseKL}.

\subsection{MuJoCo Results}

\textbf{Generalized Parametric Policies. }
By fixing the policy ratio to be $\regPi(a|s) / \pi_{\theta}(a|s)$, S$f$-AC avoids the numerical issue when $\regPi(a|s)=0$ due to the $q$-exponential.
Some papers have reported that utilizing a generalized parametric policy $\pi_{\theta}$ can significantly improve performance by better capturing the characteristics of such finite-support $\regPi$ \citep{Zhu2025-FatToThin}.
To this end, we run S$f$-AC JS with the same setting $N_\text{loss}=3$ but with a generalized parametric policy.

\begin{figure}[ht!]
    \centering
    \includegraphics[width=0.9\linewidth]{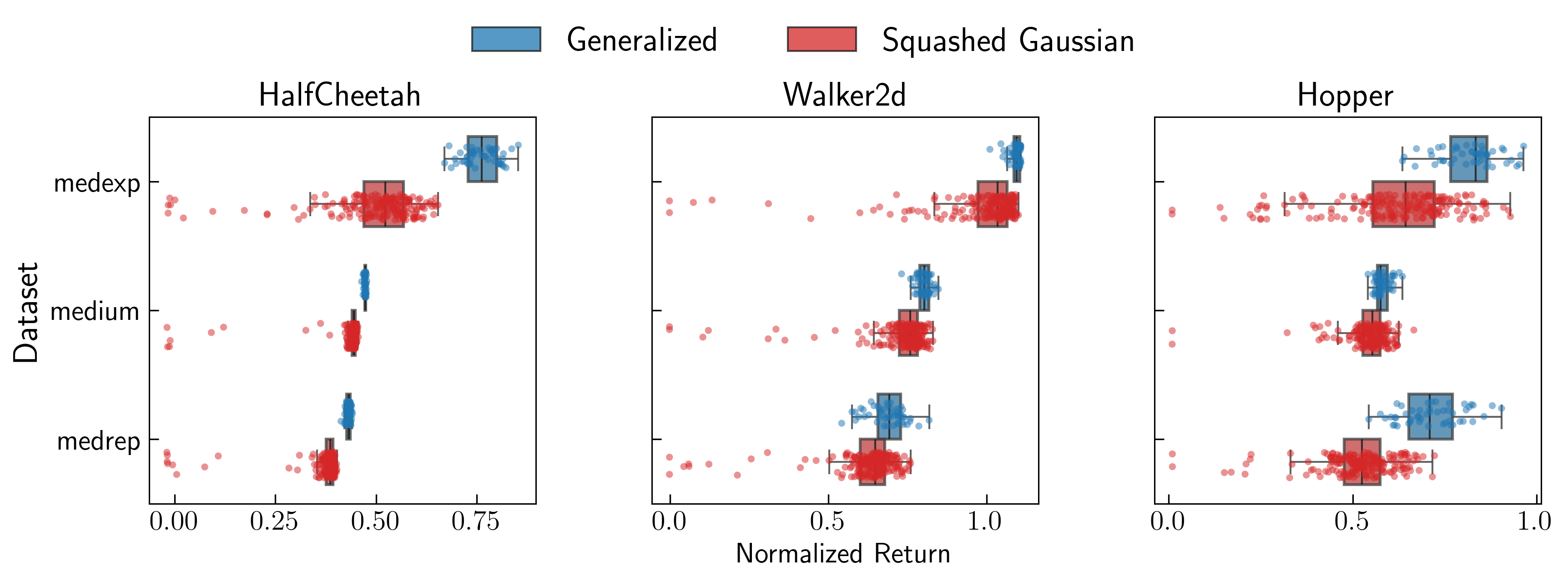}
    \caption{
    Generalized parametric policy $\pi_{\theta}$ versus the standard Squashed Gaussian policy.
    Dots are from the last $50\%$ of learning evaluation.
    Generalized $\pi_{\theta}$ can better capture the characteristics of finite-support $\regPi$ and improve S$f$-AC performance.
    }
    \label{fig:box_heavy}
\end{figure}

% \rebuttal{
Figure \ref{fig:box_heavy} compares the performance of the generalized parametric policy  against the standard Squashed Gaussian.
Dots are from the evaluation of the last $50\%$ of learning.
A generalized parametric policy significantly improves median performance across environment-dataset combinations and greatly reduces variance: low-score red dots do not appear for the generalized policy.

\begin{table}[h!]
\caption{
Averaged wallclock time (minutes) for S$f$-AC across $N$ and baselines.
}
\label{tab:computation_time}
\centering
\resizebox{0.6\columnwidth}{!}{
\begin{tabular}{cccc}
\toprule
 \textbf{JS} $N=2$ & \textbf{JS} $N=3$ & \textbf{JS} $N=5$ & \textbf{JS} $N=6$ \\
\cmidrule(lr){1-4}
% \multirow[c]{4}{*}{HalfCheetah} & \multirow[c]{4}{*}{medexp} & 364.91 $\pm$ 2.95 & 365.89 $\pm$ 3.42 & 329.57 $\pm$ 1.44 & 335.03 $\pm$ 1.61 &  \\
364.91 $\pm$ 2.95 & 365.89 $\pm$ 3.42 & 329.57 $\pm$ 1.44 & 333.67 $\pm$ 2.77 \\
\midrule
\textbf{XQL} & \textbf{IQL} & \textbf{SQL} & \textbf{BPR} \\
\cmidrule(lr){1-4}
326.80 $\pm$ 2.01 & 238.53 $\pm$ 2.18 & 227.67 $\pm$ 1.72 & 1175.09 $\pm$ 36.43 \\
 % & medium & 364.33 $\pm$ 3.22 & 359.85 $\pm$ 0.40 \\
 % & medrep & 340.89 $\pm$ 2.95 & 338.85 $\pm$ 2.77 \\
%  \cmidrule(lr){1-4}
% \multirow[c]{3}{*}{Hopper} & medexp & 405.03 $\pm$ 1.48 & 406.07 $\pm$ 1.40 \\
%  & medium & 418.54 $\pm$ 2.43 & 419.15 $\pm$ 1.99 \\
%  & medrep & 368.72 $\pm$ 5.05 & 366.95 $\pm$ 3.54 \\
%  \cmidrule(lr){1-4}
% \multirow[c]{3}{*}{Walker2d} & medexp & 369.85 $\pm$ 2.44 & 369.66 $\pm$ 4.19 \\
%  & medium & 369.44 $\pm$ 2.76 & 369.02 $\pm$ 3.51 \\
%  & medrep & 346.27 $\pm$ 2.75 & 346.50 $\pm$ 4.18 \\
 \bottomrule
\end{tabular}
}
\end{table}

\textbf{Wallclock Time Comparison. }
Table \ref{tab:computation_time} lists the wallclock time of S$f$-AC across the number of terms and against the baselines.
Increasing the number of terms does not noticeably increase computation time, and S$f$-AC remains on the same order of magnitude as XQL, though it takes slightly more time than IQL and SQL.
The computation of S$f$-AC does not require storing intermediate results or variables and hence no extra memory is required.
The following code snippet computes the series for S$f$-AC Jensen-Shannon, and it is clear that only the resulting sum is needed.
All wallclock times are recorded for 1 million steps using an Intel 8457C CPU and an Nvidia A6000 GPU.

\vspace{2mm}
\begin{lstlisting}[language=Python, caption=The series representation of the conditional symmetry term of Jensen-Shannon divergence.]
res_series = torch.sum(torch.hstack([(-1)**n / (2**(n-1) * n * (n-1)) * self.clamp_ratio((res_ratio - 1)**n) for n in range(2, self.num_terms)]), dim=1, keepdim=True)
\end{lstlisting}

\section{Implementation Details}\label{apdx:implementation}

% Preamble:
% \usepackage{booktabs}

% Preamble:
% \usepackage{booktabs}

% Preamble:
% \usepackage{booktabs}
% \usepackage{graphicx}

\begin{table}[t]
\centering
\scriptsize

\resizebox{0.99\textwidth}{!}{%
\begin{tabular}{lccccccccc}
\toprule
& \multicolumn{3}{c}{HalfCheetah} & \multicolumn{3}{c}{Hopper} & \multicolumn{3}{c}{Walker2d} \\
\cmidrule(lr){2-4} \cmidrule(lr){5-7} \cmidrule(lr){8-10}
& Medium-Expert & Medium & Medium-Replay & Medium-Expert & Medium & Medium-Replay & Medium-Expert & Medium & Medium-Replay \\
\midrule
Temperature
& $10^{-2}$ & $10^{-3}$ & $10^{-1}$
& $10^{-2}$ & $10^{-3}$ & $10^{-4}$
& $10^{-3}$ & $10^{-1}$ & $10^{-4}$ \\
Learning rate
& $10^{-3}$ & $10^{-3}$ & $10^{-4}$
& $10^{-4}$ & $3\times10^{-3}$ & $10^{-4}$
& $10^{-4}$ & $3\times10^{-4}$ & $3\times10^{-4}$ \\
\bottomrule
\end{tabular}%
}

\vspace{0.6em}

\resizebox{0.99\textwidth}{!}{%
\begin{tabular}{lccccccc}
\toprule
& Maze2D-Medium & Maze2D-Umaze & Maze2D-Large & Door-Expert & Pen-Expert & Relocate-Expert & Hammer-Expert \\
\midrule
Temperature
& $10^{-4}$ & $10^{-3}$ & $10^{-4}$ & $1$ & $10^{-1}$ & $1$ & $10^{-3}$ \\
Learning rate
& $10^{-4}$ & $3\times10^{-4}$ & $3\times10^{-4}$ & $10^{-3}$ & $10^{-4}$ & $10^{-4}$ & $10^{-4}$ \\
\bottomrule
\end{tabular}%
}
\vspace{3mm}
\caption{
The best hyperparameters of S$f$-AC Jensen-Shannon for each environment.
Selected based on final scores, averaged over 5 seeds.
}
\label{tab:fac-hparams}
\end{table}

% \subsection{Implementation Details}\label{apdx:implementation}

 % \begin{table}[t]
 %     \centering
 %     \begin{small}
 %     \resizebox{.8\textwidth}{!}{%
 %     \begin{tabular}{l cccccc}
 %         \toprule
 %         Dataset & S$f$-AC JS & S$f$-AC Jeffreys & AWAC & IQL & XQL & SQL \\
 %         \midrule
 %         HalfCheetah-Medium-Expert & $0.01$ & $0.01$ & $1.00$ & $0.33$ & $2.00$ & $5.00$ \\
 %         HalfCheetah-Medium-Replay & $0.01$ & $0.01$ & $1.00$ & $0.33$ & $2.00$ & $5.00$ \\
 %         HalfCheetah-Medium & $0.01$ & $0.01$ & $0.50$ & $0.33$ & $2.00$ & $5.00$ \\
 %         Hopper-Medium-Expert & $0.01$ & $0.01$ & $1.00$ & $0.33$ & $2.00$ & $2.00$ \\
 %         Hopper-Medium-Replay & $0.01$ & $0.01$ & $0.50$ & $0.33$ & $2.00$ & $2.00$ \\
 %         Hopper-Medium & $0.01$ & $0.01$ & $0.50$ & $0.33$ & $2.00$ & $5.00$ \\
 %         Walker2d-Medium-Expert & $0.01$ & $0.01$ & $0.10$ & $0.33$ & $2.00$ & $5.00$ \\
 %         Walker2d-Medium-Replay & $0.01$ & $0.01$ & $0.10$ & $0.33$ & $2.00$ & $5.00$ \\
 %         Walker2d-Medium & $0.01$ & $0.01$ & $0.10$ & $0.33$ & $2.00$ & $5.00$ \\
 %         \bottomrule
 %     \end{tabular}
 %     }
 %     \vspace{0.5em}
 %          \caption{
 %          The best  $\tau$ across environments.
 %            Published settings were used for baselines.
 %          }
 %     \label{tab:all_tau}
 %     \end{small}
 % \end{table}

\begin{table}[t] 
\begin{small}
\begin{center}
\resizebox{.8\textwidth}{!}{%
\begin{tabular}{lcccccc}
\toprule
Hyperparameter & Value \\
\midrule
Learning rate & \makecell{Swept in $\{3\times 10^{-3}, 1\times 10^{-3}, 3\times 10^{-4}, 1\times 10^{-4}\}$ \\See the best setting in Table \ref{tab:fac-hparams} } \\[1ex] \hline

Temperature & \makecell{Same as the number reported in \\the publication of each algorithm.\\
Swept in $\{10^{-4}, 10^{-3}, 10^{-2}, 10^{-1}, 1.0\}$\\See the best setting in Table \ref{tab:fac-hparams} \\} \\[1ex] \hline

IQL Expectile & 0.7 \\ \hline
Discount rate & 0.99 \\ \hline
Hidden size of Value network & 256 \\ \hline
Hidden layers of Value network & 2 \\ \hline
Hidden size of Policy network & 256 \\ \hline
Hidden layers of Policy network & 2 \\ \hline
Minibatch size & 256 \\ \hline
Adam.$\beta_1$ & 0.9 \\ \hline
Adam.$\beta_2$ & 0.99 \\ \hline
Number of seeds for sweeping & 5 \\ \hline
Number of seeds for the best setting & 10 \\
\bottomrule
\end{tabular}
}
\end{center}
\end{small}
\caption{Default hyperparameters and sweeping choices.}
\label{table:offline_default_param}
\end{table}

We use the D4RL benchmark \citep{Fu2020-d4rl} for offline experiments.
For MuJoCo locomotion tasks, the offline datasets each contain 1 million samples generated by a partially trained SAC agent. The dataset name reflects the training level of the agent used to collect the transitions. The Medium dataset contains samples generated by a medium-level (half-trained) SAC policy. 
Medium-expert mixes trajectories from the Medium dataset with those produced by an expert agent. Medium-replay consists of samples from the replay buffer collected during training until the policy reaches medium-level performance. In summary, the ranking of levels is Medium-expert $>$ Medium $>$ Medium-replay.
Each agent was trained for $1\times 10^6$ steps. The policy was evaluated every $1000$ steps. The score was averaged over $5$ rollouts in the real environment; each had $1000$ steps.

The Adroit datasets consist of high-dimensional dexterous manipulation tasks involving a 24-DoF robotic hand.
These datasets are challenging because of the high-dimensional continuous-control problem and narrow data support. 
The Maze2D datasets focus on navigation, with difficulty increasing with maze complexity. 
Maze2D is challenging because it requires trajectory stitching and long-horizon planning under sparse rewards. 

\textbf{Agents. }
For MuJoCo locomotion tasks, we include BPR \citep{Srinivasan-2025-BPR} that performs behavior preference regression,
AWAC, which corresponds to explicit KL regularization \citep{Nair2021-awac}, XQL, which corresponds to implicit KL regularization \citep{Garg2023-extremeQlearning}, SQL \citep{Xu2023-OfflineNoODDAlphaDiv}, which corresponds to sparse $\alpha$-divergence regularization, and IQL \citep{Kostrikov2022-implicitQlearning}.

For Adroit/Maze2D, we compare against IQL \citep{Kostrikov2022-implicitQlearning}, CQL \citep{Kumar2020-CQL}, EDAC \citep{An-2021-EDAC-uncertainty-Q-ensemble}, Decision Transformer (DT) \citep{Chen2021-decisionTransformer}, ReBRAC \citep{Tarasov-2023-ReBRAC-revisiting-minimalistOffline}.
The baseline scores are taken from the official CORL benchmark site\footnote{\url{https://corl-team.github.io/CORL/benchmarks/offline}}.

\textbf{Parameter sweeping:} 
S$f$-AC results in the paper were generated by the best parameter setting after sweeping.
For the baselines, their published settings were used.
The best learning rates and temperatures for each environment are listed in Table \ref{tab:fac-hparams}.

\textbf{Computation Overhead: }
All experiments were run with 96 Intel 8457C CPU cores.
In terms of computation time, S$f$-AC Jensen-Shannon took approximately 4 hours on average for 1 million steps.

\newpage
\section*{NeurIPS Paper Checklist}

\begin{enumerate}

\item {\bf Claims}
    \item[] Question: Do the main claims made in the abstract and introduction accurately reflect the paper's contributions and scope?
    \item[] Answer: \answerYes{} % Replace by \answerYes{}, \answerNo{}, or \answerNA{}.
    \item[] Justification: The abstract and introduction have clearly indicated our contributions.

\item {\bf Limitations}
    \item[] Question: Does the paper discuss the limitations of the work performed by the authors?
    \item[] Answer: \answerYes{} % Replace by \answerYes{}, \answerNo{}, or \answerNA{}.
    \item[] Justification: The conclusion section discusses the limitations.

\item {\bf Theory assumptions and proofs}
    \item[] Question: For each theoretical result, does the paper provide the full set of assumptions and a complete (and correct) proof?
    \item[] Answer: \answerYes{} % Replace by \answerYes{}, \answerNo{}, or \answerNA{}.
    \item[] Justification: We provide the proofs of all theorems in Appendix \ref{apdx:math_details}.

    \item {\bf Experimental result reproducibility}
    \item[] Question: Does the paper fully disclose all the information needed to reproduce the main experimental results of the paper to the extent that it affects the main claims and/or conclusions of the paper (regardless of whether the code and data are provided or not)?
    \item[] Answer: \answerYes{} % Replace by \answerYes{}, \answerNo{}, or \answerNA{}.
    \item[] Justification: We provide experiment details in Appendix \ref{apdx:implementation} and full code in supplementary material.

\item {\bf Open access to data and code}
    \item[] Question: Does the paper provide open access to the data and code, with sufficient instructions to faithfully reproduce the main experimental results, as described in supplemental material?
    \item[] Answer: \answerYes{} % Replace by \answerYes{}, \answerNo{}, or \answerNA{}.
    \item[] Justification: We provide source code in the supplementary material to allow faithful reproduction of the results.

\item {\bf Experimental setting/details}
    \item[] Question: Does the paper specify all the training and test details (e.g., data splits, hyperparameters, how they were chosen, type of optimizer) necessary to understand the results?
    \item[] Answer: \answerYes{} % Replace by \answerYes{}, \answerNo{}, or \answerNA{}.
    \item[] Justification: Appendix \ref{apdx:implementation} contains training details.

\item {\bf Experiment statistical significance}
    \item[] Question: Does the paper report error bars suitably and correctly defined or other appropriate information about the statistical significance of the experiments?
    \item[] Answer: \answerYes{} % Replace by \answerYes{}, \answerNo{}, or \answerNA{}.
    \item[] Justification: All figures and tables in the paper showing scores come with statistical significance.

\item {\bf Experiments compute resources}
    \item[] Question: For each experiment, does the paper provide sufficient information on the computer resources (type of compute workers, memory, time of execution) needed to reproduce the experiments?
    \item[] Answer: \answerYes{} % Replace by \answerYes{}, \answerNo{}, or \answerNA{}.
    \item[] Justification: Appendix \ref{apdx:implementation} provides computation overhead.
    
\item {\bf Code of ethics}
    \item[] Question: Does the research conducted in the paper conform, in every respect, with the NeurIPS Code of Ethics \url{https://neurips.cc/public/EthicsGuidelines}?
    \item[] Answer: \answerYes{} % Replace by \answerYes{}, \answerNo{}, or \answerNA{}.
    \item[] Justification: We confirm that the paper conforms with NeurIPS Code of Ethics.

\item {\bf Broader impacts}
    \item[] Question: Does the paper discuss both potential positive societal impacts and negative societal impacts of the work performed?
    \item[] Answer: \answerNA{} % Replace by \answerYes{}, \answerNo{}, or \answerNA{}.
    \item[] Justification: This paper studies reinforcement learning algorithms. There is no societal impact needs to be discussed.
    
\item {\bf Safeguards}
    \item[] Question: Does the paper describe safeguards that have been put in place for responsible release of data or models that have a high risk for misuse (e.g., pre-trained language models, image generators, or scraped datasets)?
    \item[] Answer: \answerNA{} % Replace by \answerYes{}, \answerNo{}, or \answerNA{}.
    \item[] Justification: The paper poses no such risks.

\item {\bf Licenses for existing assets}
    \item[] Question: Are the creators or original owners of assets (e.g., code, data, models), used in the paper, properly credited and are the license and terms of use explicitly mentioned and properly respected?
    \item[] Answer: \answerYes{} % Replace by \answerYes{}, \answerNo{}, or \answerNA{}.
    \item[] Justification: The sources used in the paper have all been properly cited.

\item {\bf New assets}
    \item[] Question: Are new assets introduced in the paper well documented and is the documentation provided alongside the assets?
    \item[] Answer: \answerYes{} % Replace by \answerYes{}, \answerNo{}, or \answerNA{}.
    \item[] Justification: Our supplementary material contains well-documented code.

\item {\bf Crowdsourcing and research with human subjects}
    \item[] Question: For crowdsourcing experiments and research with human subjects, does the paper include the full text of instructions given to participants and screenshots, if applicable, as well as details about compensation (if any)? 
    \item[] Answer: \answerNA{} % Replace by \answerYes{}, \answerNo{}, or \answerNA{}.
    \item[] Justification: The paper does not involve crowdsourcing nor research with human subjects.

\item {\bf Institutional review board (IRB) approvals or equivalent for research with human subjects}
    \item[] Question: Does the paper describe potential risks incurred by study participants, whether such risks were disclosed to the subjects, and whether Institutional Review Board (IRB) approvals (or an equivalent approval/review based on the requirements of your country or institution) were obtained?
    \item[] Answer: \answerNA{} % Replace by \answerYes{}, \answerNo{}, or \answerNA{}.
    \item[] Justification: The paper does not involve crowdsourcing nor research with human subjects.

\item {\bf Declaration of LLM usage}
    \item[] Question: Does the paper describe the usage of LLMs if it is an important, original, or non-standard component of the core methods in this research? Note that if the LLM is used only for writing, editing, or formatting purposes and does \emph{not} impact the core methodology, scientific rigor, or originality of the research, declaration is not required.
    %this research? 
    \item[] Answer: \answerNA{} % Replace by \answerYes{}, \answerNo{}, or \answerNA{}.
    \item[] Justification: The core method development in this research does not involve LLMs as any important, original, or non-standard components.

\end{enumerate}

\end{document}